\documentclass[sigconf, nonacm]{acmart}
\AtBeginDocument{%
}

\usepackage{amsmath,amsfonts,bm}

\def\eqref#1{equation~\ref{#1}}

\def\1{\bm{1}}

\DeclareMathAlphabet{\mathsfit}{\encodingdefault}{\sfdefault}{m}{sl}
\SetMathAlphabet{\mathsfit}{bold}{\encodingdefault}{\sfdefault}{bx}{n}

\DeclareMathOperator*{\argmin}{arg\,min}

\usepackage{booktabs}
\usepackage{multirow}
\usepackage{tabularx}
\usepackage{array}
\usepackage{caption}
\usepackage{xcolor}
\usepackage{soul}
\usepackage{colortbl}
\usepackage{placeins}
\usepackage{algorithm}
\usepackage{algpseudocode}

\usepackage{amsmath,amssymb,amssymb}

\usepackage{tabularx,multirow,multicol}
\usepackage{url}
\usepackage{subfigure}
\usepackage{float}
\usepackage{xspace}

\setcopyright{acmcopyright}
\copyrightyear{2026} 
\acmYear{2026} 
\setcopyright{rightsretained} 
\acmConference[KDD '26]{Proceedings of the 32th ACM SIGKDD Conference on Knowledge Discovery and Data Mining}{August 9--13, 2026}{Jeju, Korea}
\acmBooktitle{Proceedings of the 32th ACM SIGKDD Conference on Knowledge Discovery and Data Mining (KDD '26), August 9--13, 2026, Jeju, Korea}
\acmISBN{978-1-4503-XXXX-X/2018/06}

\begin{document}

\title{Policy4OOD: A Knowledge-Guided World Model for Policy Intervention Simulation against the Opioid Overdose Crisis}

\author{Yijun Ma*, Zehong Wang*, Weixiang Sun, Zheyuan Zhang, Kaiwen Shi, Nitesh Chawla, Yanfang Ye$^\dag$}
\affiliation{%
  \institution{University of Notre Dame, South Bend, USA}
  \city{$*$ Equal Contribution \quad $\dag$ Corresponding Author}
  \country{}
}
\email{{yma7,zwang43,wsun4,zzhang42,kshi3,nchawla,yye7}@nd.edu}



\renewcommand{\shortauthors}{Ma et al.}
\renewcommand\tabularxcolumn[1]{>{\raggedright\arraybackslash}m{#1}}

\definecolor{Gray}{gray}{0.95}
\definecolor{Blue1}{RGB}{136, 190, 220}
\definecolor{Blue2}{RGB}{218, 232, 245}
\definecolor{Blue3}{RGB}{239, 248, 253}
\definecolor{darkgreen}{RGB}{0, 100, 0}
\definecolor{darkred}{RGB}{139, 0, 0}

%


\begin{abstract}
    The opioid epidemic remains one of the most severe public health crises in the United States, yet evaluating policy interventions before implementation is difficult: multiple policies interact within a dynamic system where targeting one risk pathway may inadvertently amplify another. We argue that effective opioid policy evaluation requires three capabilities---forecasting future outcomes under current policies, counterfactual reasoning about alternative past decisions, and optimization over candidate interventions---and propose to unify them through world modeling. We introduce \textbf{\method}, a knowledge-guided spatio-temporal world model that addresses three core challenges: \textit{what} policies prescribe, \textit{where} effects manifest, and \textit{when} effects unfold.
    \method jointly encodes policy knowledge graphs, state-level spatial dependencies, and socioeconomic time series into a policy-conditioned Transformer that forecasts future opioid outcomes.
    Once trained, the world model serves as a simulator: forecasting requires only a forward pass, qualitative counterfactual reasoning substitutes alternative policy encodings in the historical sequence, and policy optimization employs Monte Carlo Tree Search over the learned simulator. To support this framework, we construct a state-level monthly dataset (2019--2024) integrating opioid mortality, socioeconomic indicators, and structured policy encodings. Experiments demonstrate that spatial dependencies and structured policy knowledge significantly improve forecasting accuracy, validating each architectural component and the potential of world modeling for data-driven public health decision support. Our code and data have been released in \url{https://github.com/antman9914/Policy4OOD}.
\end{abstract}

\keywords{World Model, Opioid Policy Intervention Simulation, Knowledge Graph, Spatio-Temporal Forecasting}

\newcommand{\method}{Policy4OOD\xspace}


\maketitle

\section{Introduction}

Opioids are a class of compounds including illicit drugs like heroin, illicitly manufactured fentanyl, and
prescription pain relievers (e.g., oxycodone) \cite{fan2018automatic}. The opioid epidemic has emerged as one of the most severe public health crises in the United States over the past two decades \cite{qian2021distilling,fan2017social}. In 2023 alone, approximately 80,000 Americans died from opioid overdoses—nearly ten times the number recorded in 1999~\cite{cdc2025,cdc_overdose_stats}. Beyond mortality, the crisis imposes substantial economic costs: opioid misuse and overdose imposed an estimated \$1.02 trillion in 2017 alone, encompassing healthcare expenditures, criminal justice costs, and lost productivity~\cite{cdc2017}. Critically, the opioid crisis cannot be attributed to a single causal pathway; rather, it emerges from a complex, evolving system of heterogeneous and interdependent mechanisms spanning prescription regulation, illicit synthetic drug markets, treatment accessibility, patient behavioral responses, and broader socioeconomic pressures~\cite{krueger2017have}. This multi-factor coupling and continual evolution render the opioid crisis a complex dynamic process that unfolds across time and space, demanding analytical frameworks capable of capturing such complexity~\cite{zhang2024dietodin,li2025interpretable,zhang2025mopihfrs,wen2022disentangle}.

Policy intervention is one of the important instruments for addressing this crisis.
Public health and policy communities have implemented a wide range of interventions targeting both supply and demand sides of opioid use, including prescription monitoring programs to restrict opioid prescribing, expanded access to medication-assisted treatment and recovery services, and naloxone distribution initiatives to prevent overdose deaths~\cite{mchugh2015prescription,clark2014systematic}. Despite their diverse objectives and mechanisms, these interventions are ultimately operationalized through concrete policy instruments—legislation, administrative regulations, funding allocations, and enforcement guidelines. 
For example, income support policy acts as a key channel where employment support improves economic stability, suppresses illicit demand, and reduces opioid overdose mortality.
In this sense, policy serves as the critical interface between intervention design and real-world outcomes: any effort to mitigate the opioid crisis must work through the policy lever.

\begin{figure*}[!t]
    \centering
    \includegraphics[width=\linewidth]{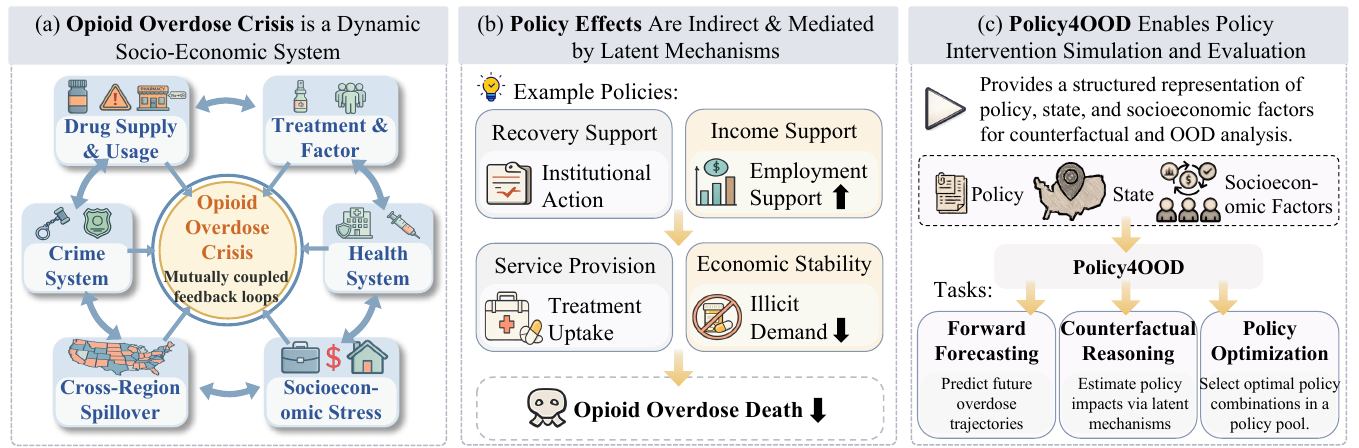}
    \caption{(a) The opioid overdose crisis emerges from a complex, dynamic socio-economic system with interacting risk factors and cross-region spillovers. (b) Policy interventions act through multiple, heterogeneous mechanisms, whose effects propagate across intermediate factors and regions. (c) Policy4OOD formulates opioid policy evaluation as a world modeling problem, enabling policy-conditioned forecasting, qualitative counterfactual analysis, and policy optimization within a unified framework.}
    \label{fig:motivation}
\end{figure*}

However, evaluating opioid-related policies remains profoundly challenging. The real-world effects of policy interventions are often difficult to anticipate prior to implementation. Multiple policies are frequently implemented concurrently, and their interactions can reshape underlying risk dynamics in ways that make the marginal effect of any single policy difficult to isolate~\cite{lee2021systematic,cerda2025role}. More fundamentally, policies do not act upon stable targets; they operate within a dynamic system where interventions targeting one pathway may induce substitution toward others. For instance, restrictive prescribing policies have been associated with reduced prescription-related harms but also with increased exposure to illicit opioids carrying higher overdose risk~\cite{wen2019prescription}. Existing evaluation approaches—whether causal inference methods estimating average treatment effects from historical data or mathematical models simulating predefined scenarios~\cite{bobashev2018pain,cerda2021systematic,lim2022modeling}—primarily focus on fitting historical trajectories and explaining past patterns. They lack the capacity to characterize how opioid-related risks may evolve under alternative policy configurations or future scenarios that have not yet been realized.

We argue that an effective policy intervention simulation framework for the opioid crisis should support three fundamental capabilities. First, \textbf{forecasting}: given the current socioeconomic conditions and a specified policy configuration, the framework should predict future opioid-related outcomes. This enables policymakers to anticipate the consequences of proposed interventions before implementation. Second, \textbf{counterfactual reasoning}: the framework should assess what would have happened had alternative policies been enacted. This capability supports retrospective evaluation of policy decisions and identification of missed opportunities. Third, \textbf{policy optimization}: given a pool of available policy instruments, the framework should identify the combination most likely to minimize adverse outcomes. This moves beyond evaluating individual policies in isolation toward systematic comparison of intervention portfolios. Together, these three capabilities transform policy evaluation from a retrospective exercise into a prospective decision-support tool.

These desiderata naturally motivate reframing opioid policy intervention simulation through the lens of \textbf{world modeling}. A world model is a learned simulator that captures the dynamics of a system and can generate trajectories under specified conditions~\cite{ha2018world}. In our context, the objective is to construct a model that forward-simulates the evolution of opioid-related outcomes conditioned on policy interventions. Once trained, such a world model directly enables all three capabilities: forecasting corresponds to forward simulation under enacted policies; qualitative counterfactual reasoning corresponds to simulation under hypothetical alternative policies; and policy optimization corresponds to searching over the space of possible policy sequences to identify those that minimize predicted adverse outcomes. This unified framework provides a principled foundation for comprehensive policy intervention simulation.

However, building such a world model for the opioid crisis presents several fundamental challenges, which we organize along three dimensions.
\textit{What} interventions do policies enact? Policy documents prescribe complex, multi-faceted intervention pathways—from regulatory provisions to behavioral changes to health outcomes—and multiple policies are often deployed simultaneously with interacting effects. Representing this structured policy knowledge in a form amenable to neural modeling is non-trivial.
\textit{Where} do policy effects manifest? A state's opioid outcomes are influenced not only by its own policies but also by those of neighboring states through cross-border spillover of drug supply, population mobility, and treatment access~\cite{dellavigna2025policy}. Models that treat regions as independent units fail to capture these spatial dependencies.
\textit{When} do policy effects take hold? The same policy may have different impacts at different stages of the crisis—immediate enforcement effects versus long-term structural changes—and the relevance of intervention strategies evolves with changing socioeconomic conditions. Capturing these temporal dynamics requires moving beyond static or linear response assumptions.

Realizing this vision requires not only methodological innovation but also suitable data infrastructure. Existing opioid-related datasets are fragmented: mortality statistics, socioeconomic indicators, and policy records reside in separate sources with inconsistent temporal granularity and geographical coverage. To bridge this gap, we construct a comprehensive state-level monthly dataset spanning 2019--2024 that integrates opioid overdose mortality from CDC WONDER, 12 socioeconomic descriptors covering economic conditions, demographic structure, public health metrics, and crime statistics, and structured policy encodings derived from substance use disorder treatment legislation. This unified dataset enables joint modeling of policy interventions and their socioeconomic context at a temporal resolution sufficient for capturing policy dynamics.

Building on this data foundation, we propose \textbf{\method}, a knowledge-conditioned spatio-temporal world model that addresses the aforementioned challenges through dedicated architectural components. To capture \textit{what} policies prescribe, we construct a policy knowledge graph from legislative documents using large language models and encode it via a relational graph neural network; a vector-quantized codebook discovers canonical intervention strategies shared across states and time periods. To model \textit{where} effects manifest, a spatial graph neural network over state adjacency captures cross-border spillover through iterative neighborhood aggregation. To learn \textit{when} effects unfold, a Transformer with dual-branch training—combining masked token prediction for robust representations and next-token prediction for causal dynamics—models temporal evolution. The trained world model serves as a fast simulator: forecasting under enacted policies requires only forward passes, qualitative counterfactual analysis substitutes alternative policy encodings, and policy optimization employs Monte Carlo Tree Search (MCTS) to efficiently navigate the combinatorial space of intervention schedules.

\section{Related Work}
\label{sec:literature}

\subsection{Policy Evaluation for Opioid Overdose Crisis}

Evaluating opioid policies is challenging due to the absence of controlled experiments, regional heterogeneity, and interactions among concurrent interventions.
\textit{(1) Causal inference approaches} estimate policy effects via quasi-experimental designs~\cite{lee2021systematic,cerda2025role}, but are inherently retrospective and struggle to isolate marginal effects when multiple policies co-occur.
\textit{(2) Simulation-based approaches} enable counterfactual analysis through system dynamics and agent-based models~\cite{cerda2021systematic,lim2022modeling,bobashev2018pain,shojaati2023abm}, yet rely on predefined behavioral assumptions and represent policies through coarse indicators.
\textit{(3) Data-driven forecasting} methods such as CASTNet~\cite{ertugrul2019castnet} and TrOP~\cite{matero2023trop} focus on prediction accuracy rather than policy-conditional simulation, lacking mechanisms to evaluate unobserved interventions~\cite{heuton2024spatiotemporal}.
A common limitation is the inability to support prospective decision-making: forecasting under proposed policies, reasoning about counterfactuals, and optimizing over possible interventions. Our work addresses this gap by reframing policy simulation as a world modeling problem, which evaluates the policies from multiple aspects.

\subsection{World Model}

World models are learned simulators that generate future trajectories conditioned on actions~\cite{ha2018world}. In reinforcement learning, the Dreamer family~\cite{hafner2019dream,hafner2020dreamerv2,hafner2023dreamerv3} and MuZero~\cite{schrittwieser2020mastering} demonstrate that learned dynamics models can substitute for hand-crafted simulators. Recent work extends world models to video prediction~\cite{yan2021videogpt,bruce2024genie}, autonomous driving~\cite{hu2023gaia}, and robotics~\cite{wu2023daydreamer}, suggesting their potential as general-purpose simulators for complex real-world dynamics.
Despite this success, world models remain unexplored in public health and policy domains. The opioid crisis presents a natural application: policy interventions serve as ``actions'' influencing population health ``states.'' Our work introduces world models to opioid policy simulation, enabling forecasting, counterfactual reasoning, and policy optimization within a unified framework.

\subsection{Spatial-Temporal Forecasting}
Spatio-temporal forecasting provides the technical foundation for modeling opioid outcomes across time and space.
\textit{(1) Multivariate time series models} have advanced from recurrent architectures~\cite{salinas2020deepar} to Transformer variants with sparse attention~\cite{zhou2021informer}, series decomposition~\cite{wu2021autoformer}, patch-based representations~\cite{nie2022time}, and exogenous variable integration~\cite{lim2021temporal,wang2024timexer}. However, these models treat covariates generically without structured policy conditioning.
\textit{(2) Spatio-temporal graph learning} models interactions among spatial entities via graph structures~\cite{jin2024survey}, with methods ranging from diffusion convolution~\cite{li2017diffusion} and temporal convolutions~\cite{yu2017spatio} to adaptive graph learning~\cite{wu2019graph} and pattern-centric graph learning~\cite{wangbeyond,wang2025generative,ma2026tgpm}. Recent work addresses causal challenges in spatio-temporal settings~\cite{xia2023cast} and applies GNNs to epidemic forecasting~\cite{liu2024review,han2025epidemiology}, naturally capturing cross-region spillover effects.
While these models excel at capturing complex dependencies, they are predominantly designed for infrastructure systems and rarely incorporate policy interventions as first-class inputs. \method bridges this gap by integrating spatio-temporal architectures to simulate how structured interventions propagate through space and time.

\begin{figure}[!t]
  \centering
  \includegraphics[width=\linewidth]{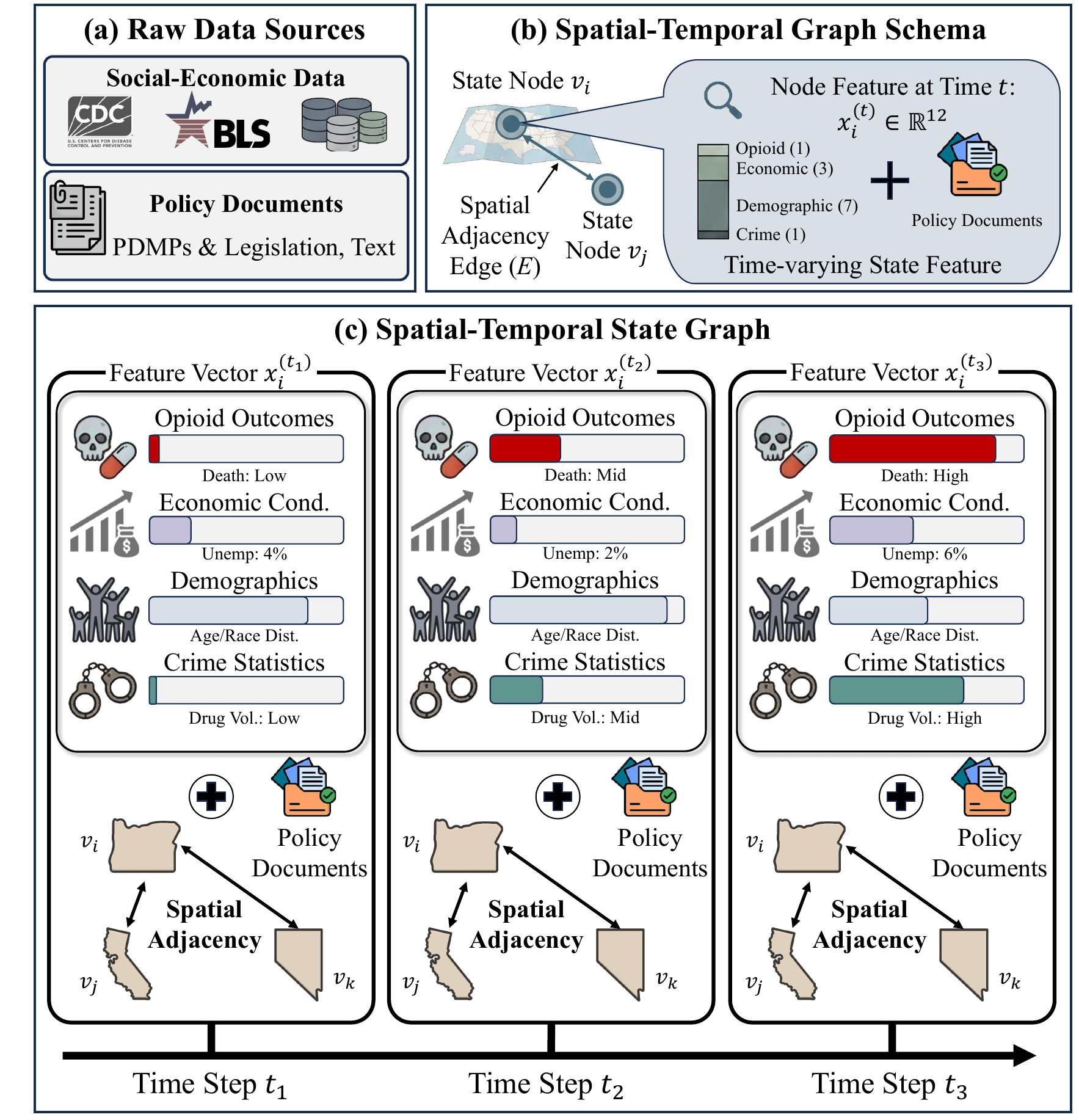}
  \caption{The schema of the spatial-temporal state graph.}
  \label{fig:dataset}
  \vspace{-5pt}
\end{figure}

\begin{figure*}[!t]
  \centering
  \includegraphics[width=\linewidth]{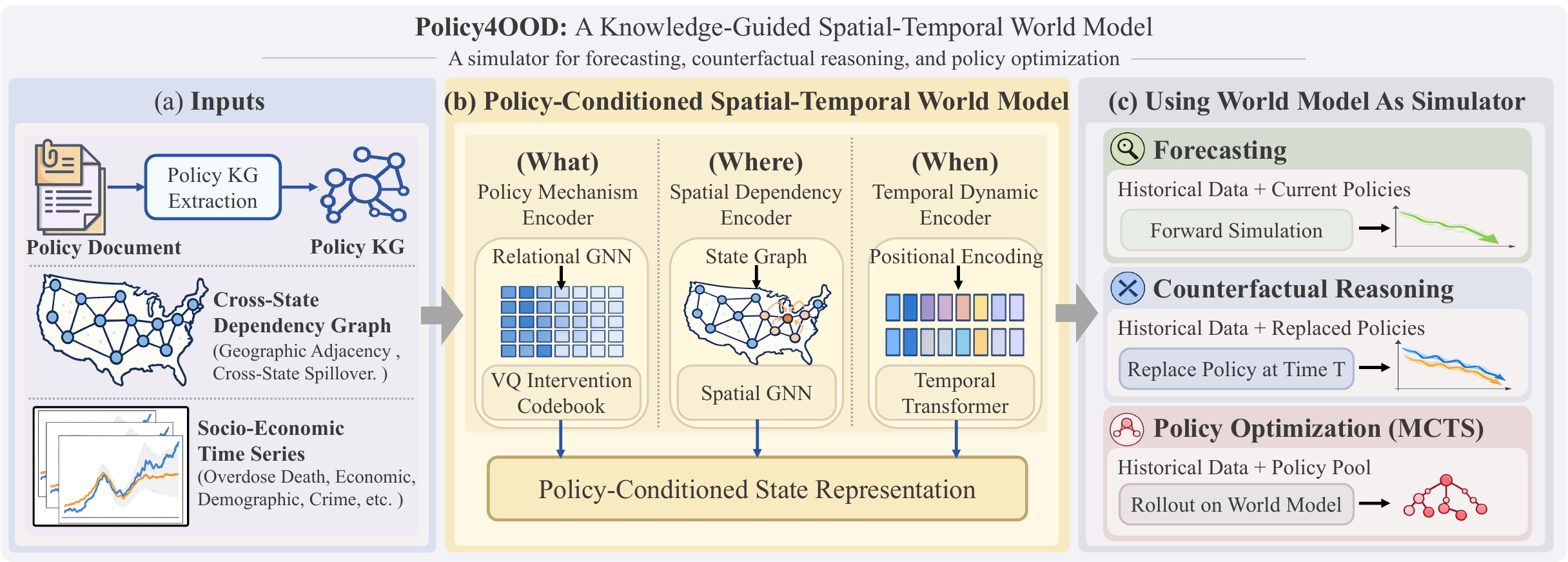}
  \caption{Overview of Policy4OOD.
    (a) Policy documents are converted into a policy knowledge graph, together with state-level socioeconomic time series and a cross-state dependency graph.
    (b) The model jointly encodes policy mechanisms (what), spatial dependencies and spillover effects (where), and temporal dynamics of policy impacts (when), producing a unified policy-conditioned state representation.
    (c) The learned world model supports forward forecasting, qualitative counterfactual reasoning through policy replacement, and policy optimization via MCTS.
  }
  \label{fig:method}
\end{figure*}

\section{Dataset Construction}
\label{sec:data}

Opioid overdose risks and policy effects are deeply intertwined with socio-economic conditions~\cite{krueger2017have,phillips2017pain}, yet existing simulation approaches often treat policies in isolation from their structural context. To enable context-aware policy modeling, we construct a spatial-temporal state graph $\mathcal{G}_S = (V, E)$ that jointly captures state-level socio-economic dynamics and inter-state dependencies. The dataset spans January 2019 to December 2024 at monthly resolution. Dataset schema is illustrated in Figure \ref{fig:dataset}. 

For graph construction, each node $v_i \in V$ represents one of the 48 contiguous U.S. states (Alaska and Hawaii are excluded due to geographical isolation). We associate each node with a time-varying feature vector $\mathbf{x}_i^{(t)} \in \mathbb{R}^{C}$ comprising $C = 12$ socio-economic descriptors.
Table~\ref{tab:descriptor} provides detailed descriptor definitions and data sources. 
These descriptors span four primary domains: opioid overdose deaths, economic indicators, demographic characteristics, and crime data. The economic indicators include unemployment rate, labor force participation rate, and unemployment insurance claims, which collectively reflect the economic health of a region. For demographic characteristics, we consider age distribution across three groups (0-18, 18-55, and 55+ years) and racial distribution (White, Black, Asian, and American Indian populations). Additionally, the number of drug/narcotic violation offenses serves as a crime indicator within our analytical framework. All data are sourced from authoritative public databases, including the Centers for Disease Control and Prevention (CDC), Federal Reserve Economic Data (FRED), StatsAmerica, and the FBI Crime Data Explorer, ensuring data reliability and traceability.
These descriptors are selected based on their established relevance in opioid policy literature and availability from authoritative national sources.

Edges $E$ encode geographical adjacency: two states are connected if they share a terrestrial border (point-contact excluded). This structure captures cross-border spillover effects---including illicit drug supply networks, population mobility, and treatment access patterns---that have been shown to influence regional opioid dynamics~\cite{dellavigna2025policy}. The resulting graph contains $|E| = 194$ edges.
For each state and time step, we collect enacted policy documents spanning prescription drug monitoring programs (PDMPs) and recovery support legislation. Raw policy text is transformed into structured representations.
Together, the socio-economic features and policy representations provide the foundation for policy-conditioned simulation.
To our knowledge, this constitutes the first integrated spatial-temporal dataset that pairs fine-grained legislative text with multi-domain socio-economic time series for data-driven opioid policy simulation at the state level.

Note that although the fact that our selected time period falls during the COVID-19 pandemic may lead to unusual opioid dynamics, the pandemic period represents a particularly high-stakes evaluation context~\cite{gomes2023trends}, where we aim to demonstrate that Policy4OOD can capture policy effects during acute socioeconomic disruption.

\begin{table}[!h]
\centering
\caption{Selected Socio-Economic Descriptors}
\begin{tabular}{p{1.6cm} p{2.5cm} p{3.2cm}}
\toprule
\textbf{Field} & \textbf{Descriptor} & \textbf{Description} \\
\midrule

Opioid 
& Overdose Death\footnotemark[1] 
& Number of opioid overdose deaths. \\
\midrule

Economic 
& Unemployment Rate\footnotemark[2] 
& Percentage of unemployed individuals. \\
\cmidrule(lr){2-3}
& Labor Force Participation\footnotemark[3] 
& Share of working-age population in labor force. \\
\cmidrule(lr){2-3}
& UI Claim\footnotemark[4] 
& Number of unemployment insurance claims. \\

\midrule

Demographic 
& Age Distribution\footnotemark[5] 
& Population aged 0--18 / 18--55 / 55+. \\
\cmidrule(lr){2-3}
& Racial Distribution\footnotemark[6] 
& Population by race (White, Black, Asian, AI). \\

\midrule

Crime 
& Drug/Narcotic Violation\footnotemark[7] 
& Number of drug-related offenses. \\

\bottomrule
\end{tabular}
\label{tab:descriptor}
\end{table}

\footnotetext[1]{\url{https://www.cdc.gov/nchs/nvss/vsrr/drug-overdose-data.htm}}
\footnotetext[2]{\url{https://fred.stlouisfed.org/series/ALUR}}
\footnotetext[3]{\url{https://fred.stlouisfed.org/series/LBSNSA01}}
\footnotetext[4]{\url{https://fred.stlouisfed.org/series/CACCLAIMS}}
\footnotetext[5]{\url{https://www.statsamerica.org/downloads/Population-by-Age-and-Sex.zip}}
\footnotetext[6]{\url{https://www.statsamerica.org/downloads/Population-by-Race.zip}}
\footnotetext[7]{\url{https://cde.ucr.cjis.gov/LATEST/webapp/\#/pages/explorer/crime/crime-trend}}

\section{Methodology}
\label{sec:method}

\subsection{Overview}
\label{sec:overview}

Modeling policy effects on the opioid crisis requires addressing three challenges: \textit{what} interventions do policies enact, \textit{where} do effects manifest, and \textit{when} do they take hold. As illustrated in Figure~\ref{fig:method}, \method addresses these through dedicated components: a policy knowledge graph with vector-quantized intervention discovery captures policy content (\textit{what}), a spatial graph neural network encodes inter-state dependencies (\textit{where}), and a Transformer with cross-attention prediction head models temporal dynamics (\textit{when}). Once trained, the world model serves as a simulator for forecasting, qualitative counterfactual analysis, and policy optimization via MCTS.

\vspace{3pt}
\noindent\textbf{Problem Definition.}
We formulate opioid policy simulation as a \textit{conditional spatio-temporal forecasting} problem. For a target state $s$, we construct a spatial graph $\mathcal{G}_s = (\mathcal{V}, \mathbf{A})$, where $\mathcal{V}$ denotes the set of neighboring states and $\mathbf{A}$ is the adjacency matrix. Given historical opioid overdose mortality $\{\mathcal{Y}^{(t)}\}_{t=1}^{T_h}$, socio-economic covariates $\{\mathbf{X}^{(t)}\}_{t=1}^{T_h}$, policy documents $\{\mathcal{P}^{(t)}\}_{t=1}^{T_h}$ with $\mathcal{P}^{(t)} = \{P_1^{(t)}, \dots, P_N^{(t)}\}$, and graph topology $\mathbf{A}$, our objective is to forecast opioid overdose mortality over the next $T_f$ months:
\begin{equation}
  P\bigl(\mathcal{Y}^{(T_h+1:T_h+T_f)} \mid \{\mathcal{Y}^{(t)}, \mathbf{X}^{(t)}, \mathcal{P}^{(t)}\}_{t=1}^{T_h}, \mathbf{A}\bigr).
\end{equation}
This graph-based formulation captures spatial dependencies, enabling predictions for each state to incorporate neighboring dynamics and cross-border policy spillover effects.

\subsection{State Representation}
\label{sec:state_repr}

Opioid dynamics exhibit strong interdependencies among geographically proximate states through drug supply networks, population mobility, and policy spillovers~\cite{dellavigna2025policy}. To capture these, we refer to graph learning methods~\cite{qian2022co,wang2024gft,zhao2023self,ju2022grape,ju2022multi} and represent each state by encoding its local spatial context. For state $s$, we extract its $K$-hop neighborhood subgraph $\mathcal{G}_s^{(K)}$ containing all states reachable within $K$ hops along geographical adjacency edges.
We encode this context through recursive neighborhood aggregation \citep{kipf2017gcn,hamilton2017inductive,velivckovic2018graph}. At each layer $k \in \{1, \dots, K\}$, the representation of the central node is updated by aggregating information from its neighbors: $\mathbf{h}_i^{(k)} = \sigma\!\left(\mathbf{W}^{(k)} \cdot \text{AGG}(\{\mathbf{h}_j^{(k-1)} : j \in \mathcal{N}(i)\})\right)$,
where $\mathcal{N}(s)$ denotes the neighbors of state $s$, $\text{AGG}(\cdot)$ is a permutation-invariant aggregation function, $\mathbf{W}^{(k)}$ are learnable parameters, and $\sigma$ is a nonlinearity.
After $K$ layers, the central node's representation $\mathbf{h}_i^{(K)}$ integrates information from its entire $K$-hop neighborhood, capturing both the state's own features and the contextual influence from proximate states.

We use this embedding as the state-level representation, computed at each time step $t$:
\begin{equation}
  \mathbf{h}_i^{t} = \mathbf{h}_i^{(K)} \big|_{\mathbf{X} = \mathbf{X}^{(t)}} \in \mathbb{R}^{d},
  \label{eq:gnn_enc}
\end{equation}
where node features are initialized with socioeconomic covariates at time $t$, i.e., $\mathbf{h}_j^{(0)} = \mathbf{W}_0 \mathbf{x}_j^{(t)}$ for all $j \in \mathcal{G}_i^{(K)}$. The aggregation parameters $\{\mathbf{W}^{(k)}\}_{k=0}^{K}$ are shared across all states and time steps.

\subsection{Policy Representation}
\label{sec:policy_repr}

Opioid policies are inherently compositional: a single legislative act may simultaneously introduce prescription limits, expand naloxone access, and fund treatment programs, each interacting with provisions from other co-enacted laws. Representing policies as flat text discards this relational structure. We therefore adopt a three-stage approach: (1) extract structured policy knowledge graphs preserving relational structure among provisions; (2) learn pathway-aware entity embeddings and discover canonical intervention strategies via vector quantization; and (3) construct a policy-conditioned state representation by softly retrieving relevant strategies from the learned vocabulary.

\vspace{3pt}
\noindent\textbf{Policy Knowledge Graph Construction.}
For state $s$ at time $t$, we collect all concurrently enacted policy documents and jointly extract structured triplets using a large language model. Constructing a single KG from co-deployed policies---rather than separate KGs per document---naturally captures cross-policy interactions: shared entities link provisions from different policies, revealing how multiple interventions jointly shape the regulatory landscape.
Formally, the policy KG is defined as $\mathcal{G}_{\text{KG}}^{(i,t)} = (\mathcal{V}_{\text{KG}}, \mathcal{E}_{\text{KG}}, \mathcal{R})$, where $\mathcal{V}_{\text{KG}}$ is the set of entity nodes, $\mathcal{E}_{\text{KG}}$ is the set of directed edges (triplets), and $\mathcal{R}$ is the set of relation types. Each triplet $(v_s, r, v_o) \in \mathcal{E}_{\text{KG}}$ with $r \in \mathcal{R}$ represents a structured relationship between a subject entity $v_s$ and an object entity $v_o$ extracted from the policy text (e.g., \textit{(prescription\_monitoring\_program, restricts, opioid\_prescriptions)}).

\vspace{3pt}
\noindent\textbf{Learning Intervention Pathway Embeddings.}
Individual policy entities rarely act in isolation; their effects depend on the broader regulatory context. To learn context-aware representations, we encode each node by propagating information along the KG structure via relational neighborhood aggregation~\cite{schlichtkrull2018modeling}.
Each entity node $v_j \in \mathcal{V}_{\text{KG}}$ is initialized with a semantic embedding from a pre-trained sentence encoder~\cite{reimers2019sentence}: $\mathbf{e}_j^{(0)} = \text{SBERT}(v_j) \in \mathbb{R}^{d_s}$. Through multi-hop aggregation over the KG, each node's representation is enriched with information from connected entities:
\begin{equation}
  \mathbf{e}_j = \text{Encode}_\Phi(\mathcal{G}_{\text{KG}}^{(s,t)}, \mathbf{E}^{(0)})_j \in \mathbb{R}^{d_p},
  \label{eq:kg_gnn}
\end{equation}
where $\mathbf{E}^{(0)} \in \mathbb{R}^{|\mathcal{V}_{\text{KG}}| \times d_s}$ is the initial embedding matrix, $\Phi$ denotes the learnable parameters, and $d_p$ is the policy embedding dimension. The resulting embedding $\mathbf{e}_j$ encodes not merely the semantics of an individual provision but the intervention pathway it participates in---capturing upstream regulatory preconditions, co-occurring provisions, and downstream policy targets.

\vspace{3pt}
\noindent\textbf{Intervention Strategy Discovery.}
Real-world policy KGs can contain thousands of nodes, making direct attention over all entity embeddings computationally prohibitive. Moreover, many provisions across different states and time periods serve similar functional roles. To discover these canonical \emph{intervention strategies}, we introduce a learnable codebook via vector quantization (VQ)~\cite{oord2017neural}.
We maintain a codebook $\mathcal{C} = \{\mathbf{c}_1, \mathbf{c}_2, \dots, \mathbf{c}_M\} \subset \mathbb{R}^{d_p}$ of $M$ learnable code vectors, each representing a canonical intervention strategy. The codebook is shared across all states and time steps.
Each entity embedding is mapped to its nearest codebook entry:
\begin{equation}
  k_j^* = \argmin_{k \in \{1,\dots,M\}} \left\| \mathbf{e}_j - \mathbf{c}_k \right\|_2.
  \label{eq:vq_argmin}
\end{equation}
To enable gradient flow through this discrete operation, we employ the straight-through estimator~\cite{oord2017neural}: the forward pass uses $\mathbf{c}_{k_j^*}$, while the backward pass passes gradients directly to $\mathbf{e}_j$:
$\mathbf{q}_j = \mathbf{e}_j + \text{sg}\left[\mathbf{c}_{k_j^*} - \mathbf{e}_j\right] \in \mathbb{R}^{d_p}$,
where $\text{sg}[\cdot]$ denotes the stop-gradient operator. Codebook vectors are updated via exponential moving average~\cite{oord2017neural}. The KG encoder is trained with a commitment loss:
\begin{equation}
  \mathcal{L}_{\text{vq}} = \frac{1}{|\mathcal{V}_{\text{KG}}|} \sum_{j=1}^{|\mathcal{V}_{\text{KG}}|} \left\| \mathbf{e}_j - \text{sg}[\mathbf{c}_{k_j^*}] \right\|_2^2,
  \label{eq:vq_loss}
\end{equation}
combined with the downstream forecasting objective, which jointly shapes the entity embeddings to be both quantization-friendly and predictive of opioid outcomes.

\vspace{3pt}
\noindent\textbf{Policy-Conditioned State Representation.}
Given the learned codebook $\mathcal{C}$, we construct the policy-conditioned representation through two parallel branches that softly retrieve relevant intervention strategies. For a query embedding $\mathbf{u}$, we compute cosine similarity against each code vector, apply softmax to obtain attention weights, and aggregate:
\begin{equation}
  \alpha_k = \frac{\exp\!\bigl(\text{cos}(\mathbf{u}, \mathbf{c}_k) / \tau\bigr)}{\sum_{k'=1}^{M} \exp\!\bigl(\text{cos}(\mathbf{u}, \mathbf{c}_{k'}) / \tau\bigr)}, \quad
  \mathcal{R}(\mathbf{u}, \mathcal{C}) = \sum_{k=1}^{M} \alpha_k \mathbf{c}_k,
  \label{eq:code_retrieve}
\end{equation}
where $\tau$ is a temperature parameter. The first branch uses the state embedding $\mathbf{h}_s^{(t)}$ to retrieve strategies relevant to the current socioeconomic context; the second uses the mean-pooled KG embedding $\bar{\mathbf{e}}^{(s,t)} = \frac{1}{|\mathcal{V}_{\text{KG}}|} \sum_{j} \mathbf{e}_j$ to retrieve strategies aligned with the enacted policies. The two retrievals are projected and concatenated:
\begin{equation}
  \mathbf{p}_s^{(t)} = \left[\mathbf{W}_1 \, \mathcal{R}(\mathbf{h}_s^{(t)}, \mathcal{C}) \;\big\|\; \mathbf{W}_2 \, \mathcal{R}(\bar{\mathbf{e}}^{(s,t)}, \mathcal{C}) \right],
  \label{eq:policy_combined}
\end{equation}
where $\mathbf{W}_1, \mathbf{W}_2$ are learnable projections and $\|$ denotes concatenation. This dual-branch design captures both which strategies are relevant given the state's conditions and which are active given the enacted policies.

\subsection{Temporal Dynamics}
\label{sec:temporal}

Policy effects are inherently temporal: the same intervention may produce immediate enforcement effects, delayed behavioral responses, and long-term structural changes. Moreover, the relevance of a given strategy evolves as the crisis transitions between epidemiological phases. Capturing these dynamics requires modeling that attends to the full historical trajectory rather than relying on fixed-lag assumptions.

\vspace{3pt}
\noindent\textbf{Policy-Conditioned Spatial-Temporal Fusion.}
We fuse the spatial state embedding with the policy-conditioned representation:
\begin{equation}
  \mathbf{z}_i^{(t)} = [\mathbf{h}_i^{(t)} \;\|\; \mathbf{W}_f \mathbf{p}_i^{(t)}],
  \label{eq:fusion}
\end{equation}
where $\mathbf{W}_f \in \mathbb{R}^{d \times d_p}$ is a learnable projection. Temporal order is injected via sinusoidal positional encodings~\cite{vaswani2017attention}:
\begin{equation}
  \hat{\mathbf{z}}_i^{(t)} = [\mathbf{z}_i^{(t)} \;\|\; \text{PE}(t)].
  \label{eq:add_pe}
\end{equation}

\vspace{3pt}
\noindent\textbf{Backbone.}
The sequence $\hat{\mathbf{Z}}_i = [\hat{\mathbf{z}}_i^{(1)}, \dots, \hat{\mathbf{z}}_i^{(T_h)}]$ is processed by a Transformer encoder~\cite{vaswani2017attention}. Through self-attention, each time step attends to the full sequence, capturing both local transitions and long-range trends. The encoder produces contextualized representations $\mathbf{O}_i = [\mathbf{o}_i^{(1)}, \dots, \mathbf{o}_i^{(T_h)}] \in \mathbb{R}^{T_h \times d}$ that serve as key-value memory for the prediction head.

\vspace{3pt}
\noindent\textbf{Prediction Head.}
Given the encoded historical context $\mathbf{O}_i$, the prediction head forecasts outcomes at each of the $T_f$ future time steps via cross-attention. For each horizon $\tau \in \{1, \dots, T_f\}$, a learnable query token $\mathbf{q}_\tau \in \mathbb{R}^{d}$ attends over the historical sequence:
\begin{equation}
  \hat{\mathbf{y}}_i^{(\tau)} = \text{CrossAttn}(\mathbf{q}_\tau, \mathbf{O}_i, \mathbf{O}_i),
  \label{eq:pred_attn}
\end{equation}
and a final projection maps to the outcome space $\hat{\mathcal{Y}}_i^{(\tau)} = \mathbf{W}_o \hat{\mathbf{y}}_i^{(\tau)} + \mathbf{b}_o$.
Since policy information is already encoded in $\mathbf{O}_i$, the prediction head requires no additional policy input. This non-autoregressive design produces all $T_f$ predictions in parallel, with each query specializing in a different forecast horizon.

\vspace{3pt}
\noindent\textbf{Training Objective.}
Training examples are constructed via a sliding window over the time series of length $T$. At each position $t$, the window $[t, t{+}T_h{-}1]$ serves as historical context and the model predicts outcomes over the subsequent $T_f$ steps, yielding $T - T_h - T_f + 1$ training instances per state. The model is jointly optimized with:
\begin{equation}
  \mathcal{L} = \mathcal{L}_{\text{pred}} + \lambda \mathcal{L}_{\text{vq}},
  \label{eq:total_loss}
\end{equation}
where $\mathcal{L}_{\text{pred}}$ is the outcome prediction loss:
\begin{equation}
  \mathcal{L}_{\text{pred}} = \frac{1}{N T_f} \sum_{i=1}^{N} \sum_{\tau=1}^{T_f} \left\| \hat{\mathcal{Y}}_i^{(\tau)} - \mathcal{Y}_i^{(\tau)} \right\|_2^2,
  \label{eq:pred_loss}
\end{equation}
and $\lambda$ controls the weight of the VQ commitment loss (Eq.~\ref{eq:vq_loss}).

\subsection{Utilizing the World Model}
\label{sec:utilize_wm}

Once trained, \method defines a differentiable simulation function $\hat{\mathcal{Y}}_i = f_\text{WM}(\{\mathbf{x}_i^{(t)}, \mathbf{p}_i^{(t)}\}_{t=1}^{T_h}, \mathbf{A})$ that maps historical inputs to future outcome trajectories $\hat{\mathcal{Y}}_i \in \mathbb{R}^{T_f \times C_O}$. Because policy information enters exclusively through the historical encoding, all downstream applications reduce to querying this function with appropriately constructed inputs.

\vspace{3pt}
\noindent\textbf{Forecasting under Enacted Policies.}
Given observed features and policy representations up to time $T_h$, a single forward pass yields predicted trajectories $\hat{\mathcal{Y}}_i^{(1)}, \dots, \hat{\mathcal{Y}}_i^{(T_f)}$, providing projections of how current conditions and existing policies will shape future opioid-related risk.

\vspace{3pt}
\noindent\textbf{Counterfactual Reasoning.}
\method supports qualitative counterfactual reasoning by modifying policy inputs within the historical sequence. A counterfactual query replaces the policy at a target step $t^*$ with an alternative $a'$, while keeping all other steps unchanged. Re-running the model produces the counterfactual trajectory $\hat{\mathcal{Y}}_i^{\text{cf}}$, and the effect is measured as:
\begin{equation}
  \Delta_i(t^*, a') = \sum_{\tau=1}^{T_f} \hat{\mathcal{Y}}_i^{(\tau)\text{cf}} - \sum_{\tau=1}^{T_f} \hat{\mathcal{Y}}_i^{(\tau)\text{fact}},
  \label{eq:cf_effect}
\end{equation}
where negative $\Delta$ indicates the alternative would have reduced adverse outcomes. This extends to multi-step interventions by replacing policies at multiple time steps.

\vspace{3pt}
\noindent\textbf{Policy Optimization via MCTS.}
For prospective policy selection, we inject a subset of candidate policy $\mathcal{A}_k \subset \{a_0, a_1, \dots, a_K\}$ at the last position of the historical sequence. The objective is to find the policy minimizing cumulative predicted adverse outcome:
\begin{equation}
  a^* = \argmin_{\mathcal{A}_k} \sum_{\tau=1}^{T_f} \hat{\mathcal{Y}}_{i}^{(\tau)}(a_k).
  \label{eq:mcts_obj}
\end{equation}
For multi-step planning over $D$ sequential decision periods, the search space grows as $\mathcal{O}((K{+}1)^{D})$, making exhaustive enumeration intractable. Given insufficient training data and discrete policy search space, we employ MCTS~\cite{browne2012survey, kocsis2006bandit} to navigate this space. Each node corresponds to a partial policy sequence, and the world model serves as a fast simulator: at each tree level, the selected policy is injected into the historical window, the model produces a trajectory, and the window is advanced. The search follows standard MCTS phases---selection via UCT~\cite{kocsis2006bandit}, expansion, rollout simulation, and backpropagation---and after $B$ simulations, the optimal sequence is extracted by following the most-visited children at each level.

\begin{table*}[!t]
  \centering
  \caption{\textbf{Results of opioid overdose death forecasting.} We evaluate under (a) standard in-distribution and (b) cross-state generalization (OOD) settings, where 40 states are used for training and 9 unseen states for testing. Results are averaged over 3 runs. The best and second-best results are in \textbf{bold} and \underline{underline}. A.R. denotes the average ranking across all metrics.}
  \vspace{-10pt}
  \small
  \begin{tabular}{l cccccc c}
    \toprule
    \textbf{Method}           & MAE$_{1\text{-}3}$       & RMSE$_{1\text{-}3}$      & MAE$_{4\text{-}6}$       & RMSE$_{4\text{-}6}$      & MAE                      & RMSE                     & A.R.$\downarrow$ \\
    \midrule
    \rowcolor{gray!12} \multicolumn{8}{l}{\textbf{(a) In-Distribution (ID)}} \\
    
    LSTM
    & \textbf{0.1271}±0.006
    & 0.2594±0.008
    & 0.2326±0.006
    & 0.4200±0.008
    & 0.1799±0.006
    & 0.3397±0.007
    & 6.00 \\

    TCN
    & 0.1281±0.005
    & \textbf{0.2319}±0.008
    & 0.2332±0.015
    & 0.4113±0.019
    & 0.1806±0.010
    & 0.3216±0.014
    & 4.83 \\

    Transformer
    & 0.1277±0.005
    & 0.2409±0.018
    & 0.2220±0.006
    & 0.4089±0.012
    & \underline{0.1749}±0.006
    & 0.3249±0.016
    & 3.58 \\

    TimesNet
    & 0.1311±0.005
    & 0.2506±0.003
    & 0.2269±0.015
    & 0.4131±0.015
    & 0.1790±0.010
    & 0.3319±0.009
    & 5.67 \\

    \cmidrule(lr){1-8}

    MTGNN
    & 0.1347±0.007
    & 0.2409±0.005
    & \underline{0.2173}±0.010
    & \underline{0.3992}±0.007
    & 0.1760±0.007
    & \underline{0.3201}±0.006
    & \underline{3.08} \\

    STGCN
    & 0.3561±0.023
    & 0.5508±0.012
    & 0.4054±0.042
    & 0.6060±0.033
    & 0.3808±0.032
    & 0.5784±0.018
    & 10.00 \\

    Graph Wavelet
    & 0.1404±0.007
    & 0.2610±0.005
    & 0.2267±0.003
    & 0.4157±0.004
    & 0.1836±0.005
    & 0.3383±0.004
    & 7.50 \\

    AGCRN
    & 0.2498±0.010
    & 0.4658±0.017
    & 0.2679±0.011
    & 0.4960±0.022
    & 0.2588±0.010
    & 0.4809±0.020
    & 9.00 \\

    STAEFormer
    & 0.1784±0.010
    & 0.3147±0.013
    & 0.2489±0.009
    & 0.4536±0.012
    & 0.2137±0.008
    & 0.3842±0.009
    & 7.50 \\

    \cmidrule(lr){1-8}

    \rowcolor{blue!5}
    \method
    & \underline{0.1273}±0.007
    & \underline{0.2361}±0.002
    & \textbf{0.2060}±0.008
    & \textbf{0.3868}±0.010
    & \textbf{0.1666}±0.002
    & \textbf{0.3114}±0.005
    & \textbf{1.33} \\

    \midrule

    \rowcolor{gray!12}
    \multicolumn{8}{l}{\textbf{(b) Out-Of-Distribution (OOD on Unseen States): 40 training / 9 testing}} \\

    LSTM
    & 0.1995±0.024
    & 0.4099±0.050
    & 0.2589±0.023
    & 0.4964±0.045
    & 0.2292±0.024
    & 0.4531±0.048
    & 6.17 \\

    TCN
    & 0.2109±0.017
    & 0.4093±0.046
    & 0.2730±0.011
    & 0.5171±0.042
    & 0.2419±0.014
    & 0.4632±0.044
    & 7.33 \\

    Transformer
    & 0.1871±0.041
    & 0.3636±0.116
    & 0.2516±0.023
    & 0.4904±0.079
    & 0.2194±0.032
    & 0.4270±0.097
    & 4.50 \\

    TimesNet
    & 0.2141±0.027
    & 0.4750±0.093
    & 0.2913±0.041
    & 0.6011±0.144
    & 0.2527±0.034
    & 0.5381±0.118
    & 8.83 \\

    \cmidrule(lr){1-8}

    MTGNN
    & \underline{0.1659}±0.024
    & \underline{0.2873}±0.066
    & \underline{0.2466}±0.023
    & \underline{0.4308}±0.040
    & \underline{0.2062}±0.023
    & \underline{0.3590}±0.053
    & \underline{2.00} \\

    STGCN
    & 0.7000±0.125
    & 0.9190±0.166
    & 0.7513±0.152
    & 1.0071±0.189
    & 0.7256±0.139
    & 0.9630±0.177
    & 10.00 \\

    Graph Wavelet
    & 0.2354±0.024
    & 0.3791±0.071
    & 0.2676±0.033
    & 0.4635±0.093
    & 0.2515±0.023
    & 0.4213±0.081
    & 5.83 \\

    AGCRN
    & 0.1802±0.033
    & 0.3251±0.045
    & 0.2638±0.023
    & 0.4557±0.034
    & 0.2220±0.028
    & 0.3904±0.040
    & 4.00 \\

    STAEFormer
    & 0.5524±0.110
    & \textbf{0.6743}±0.111
    & 0.5096±0.042
    & 0.6223±0.022
    & 0.5310±0.057
    & 0.6483±0.053
    & 8.83 \\

    \cmidrule(lr){1-8}

    \rowcolor{blue!5}
    \method
    & \textbf{0.1570}±0.062
    & \textbf{0.2702}±0.176
    & \textbf{0.2270}±0.052
    & \textbf{0.3711}±0.173
    & \textbf{0.1920}±0.057
    & \textbf{0.3206}±0.175
    & \textbf{1.00} \\

    \bottomrule
\end{tabular}
  \label{tab:main_result}
  \vspace{-10pt}
\end{table*}

\begin{table}[!t]
  \centering
  \caption{\textbf{Ablation study.} We evaluate the contribution of each component by removing or replacing it on the ID test set.}
  \vspace{-10pt}
    \begin{tabular}{l cccc}
      \toprule
      \textbf{Variant}          & RMSE$_{1\text{-}3}$ & RMSE$_{4\text{-}6}$ & MAE             & RMSE            \\
      \midrule
      \rowcolor{blue!5} \method & \textbf{0.2361}     & \textbf{0.3868}     & \textbf{0.1666} & \textbf{0.3114} \\
      \midrule
      \rowcolor{gray!12} \multicolumn{5}{l}{\textit{(a) Policy Analysis}}                                       \\
      \quad w/o VQ              & 0.2507              & 0.4049              & 0.1827          & 0.3278          \\
      \quad w/o KG Encoder      & 0.2610              & 0.4076              & 0.1799          & 0.3343          \\
      \quad w/o KG              & 0.3341              & 0.4535              & 0.2404          & 0.3938          \\
      \quad w/o Policy          & 0.2615              & 0.4115              & 0.1793          & 0.3365          \\
      \cmidrule(lr){1-5}
      \rowcolor{gray!12} \multicolumn{5}{l}{\textit{(b) Feature Analysis}}                                      \\
      \quad w/o Economic        & 0.2479              & 0.4078              & 0.1702          & 0.3279          \\
      \quad w/o Crime           & 0.2419              & 0.4157              & 0.1801          & 0.3288          \\
      \quad w/o Demographic     & 0.2626              & 0.4007              & 0.1911          & 0.3316          \\
      \cmidrule(lr){1-5}
      \rowcolor{gray!12} \multicolumn{5}{l}{\textit{(c) World Model Analysis}}                                  \\
      \quad w/o PE              & 0.2555              & 0.4172              & 0.1794          & 0.3364          \\
      \quad w/ GCN              & 0.4880              & 0.5231              & 0.3205          & 0.5055          \\
      \quad w/ GAT              & 0.4944              & 0.5697              & 0.3359          & 0.5320          \\
      \quad w/ 2-layer          & 0.2808              & 0.4198              & 0.2025          & 0.3503          \\
      \bottomrule
    \end{tabular}
  \label{tab:ablation}
  \vspace{-10pt}
\end{table}

\section{Experiments}
\label{sec:experiment}

\subsection{Experiment Setup}

\textbf{Baselines.} To demonstrate the superiority of Policy4OOD in future opioid epidemic trend prediction, we select a range of time series forecasting and spatial-temporal graph learning baselines. Time series forecasting baselines include LSTM~\cite{hochreiter1997lstm}, TCN~\cite{bai2018empirical}, Transformer~\cite{vaswani2017attention} and TimesNet~\cite{wutimesnet}. For spatial-temporal GNNs, we adopt MTGNN~\cite{wu2020mtgnn}, STGCN~\cite{stgcn}, Graph Wavelet~\cite{wu2019graph}, AGCRN~\cite{bai2020adaptive} and STAEFormer~\cite{liu2023spatio} for comparison. Baseline details are summarized as follows:

\begin{itemize}
    \item \textbf{LSTM}~\cite{hochreiter1997lstm}: A classic recurrent architecture that introduces gating mechanisms to capture long-range temporal dependencies.

    \item \textbf{TCN}~\cite{bai2018empirical}: A temporal convolutional network that applies dilated causal convolutions with residual connections, enabling parallelizable sequence modeling with large receptive fields.

    \item \textbf{Transformer}~\cite{vaswani2017attention}: The foundational self-attention architecture that models pairwise token dependencies globally at each layer, widely adopted as a strong baseline for multivariate time series forecasting.

    \item \textbf{TimesNet}~\cite{wutimesnet}: Reshapes 1D time series into 2D representations to capture multi-periodicity, then applies 2D convolutions to model both intra-period and inter-period temporal variations.

    \item \textbf{MTGNN}~\cite{wu2020mtgnn}: Jointly learns a latent graph structure from multivariate time series via a graph learning module and performs forecasting through mix-hop graph convolution interleaved with temporal convolutions.

    \item \textbf{STGCN}~\cite{stgcn}: Combines Chebyshev graph convolutions for spatial dependency modeling with gated temporal convolutions in a fully convolutional encoder-decoder framework for traffic forecasting.

    \item \textbf{Graph WaveNet}~\cite{wu2019graph}: Introduces a self-adaptive adjacency matrix learned end-to-end alongside stacked dilated causal convolutions, enabling the model to discover hidden spatial dependencies without a predefined graph.

    \item \textbf{AGCRN}~\cite{bai2020adaptive}: Learns node-specific graph structures through two adaptive modules—Node Adaptive Parameter Learning and Data Adaptive Graph Generation—and integrates them into a recurrent architecture for traffic flow prediction.

    \item \textbf{STAEformer}~\cite{liu2023spatio}: Augments a vanilla Transformer with spatio-temporal adaptive embeddings that encode node identity and time-step context, achieving state-of-the-art traffic forecasting without modifying the core attention mechanism.
\end{itemize}

\vspace{3pt}
\noindent\textbf{Evaluation Settings.} For all the methods, we use data during 2019-2022 for training, and predict overdose deaths during 2022-2023 and 2023-2024 during validation and testing respectively. Input and output window size are set as $T_h=6$ and $T_f=6$. We adopt Mean Absolute Error (MAE) and Root Mean Squared Error (RMSE)~\cite{bai2020adaptive,bai2018empirical} for evaluation. Except for the overall MAE and RMSE, we further report prediction errors over different forecasting horizons, including the first three future months (MAE$_{1-3}$,RMSE$_{1-3}$) and the last three future months (MAE$_{4-6}$,RMSE$_{4-6}$), to assess short-term and longer-term forecasting performance respectively. We consider two settings for forecasting task evaluation: (1) Standard forecasting, where models are trained to predict future trends of opioid overdose deaths based on all the available historical observations. (2) Cross-state generalization, where models are trained on data from a randomly selected subset of 40 states and evaluated on the remaining 9 unseen states. Each model is trained on 3 distinct random seeds to reduce the influence of randomness.

\vspace{3pt}
\noindent\textbf{Implementation Details.}
\texttt{GPT-4o-mini} and \texttt{all-MiniLM-L6-v2} are adopted for policy knowledge graph extraction and textual feature embedding respectively.
The default hyperparameters are: hidden dimension $d_h = 64$, projected policy dimension $d_p = 6$, number of Transformer layers $L = 1$, number of attention heads $n_{\text{head}} = 4$, feed-forward dimension $4d_h = 256$, dropout rate $0.1$
, EMA momentum coefficient $\alpha = 0.99$, learning rate $5 \times 10^{-4}$ with the Adam optimizer, and early stopping patience of 10 epochs.
Simulation budget for MCTS is set as $B=1000$.
We conduct 3 independent runs with different random seeds and report the mean value of all metrics.

\subsection{Opioid Overdose Death Forecasting}
\label{sec:main}

\begin{table*}[!htbp]
    \centering
    \caption{Policy Codes for Policy Analysis.}
    \resizebox{1.8\columnwidth}{!}{
    \begin{tabularx}{\textwidth}{lcc
    >{\hsize=0.5\hsize\raggedright\arraybackslash}X}
        \toprule
        \textbf{Policy Type}                & \textbf{Code}                    & \textbf{Attribute} & \multicolumn{1}{c}{\textbf{Description}}                                                                                                \\
        \midrule
        \multirow{6}{*}{\shortstack{Prescription\\Control}}    
                                            & prescriber\_mandatory\_PDMP\_use & Binary             & If PDMP use is mandatory for prescribers.                                                                                               \\
        \cmidrule{2-4}
                                            & dispenser\_mandatory\_PDMP\_use  & Binary             & If PDMP use is mandatory for opioid dispensers.                                                                                         \\
        \cmidrule{2-4}
                                            & substance\_monitored             & Categorical        & The coverage of monitored substances, including Schedules II - V, Schedules II - IV and Drugs of Concern.                              \\
        \cmidrule{2-4}
                                            & max\_initial\_days\_adult        & Integer            & Maximum initial prescription day for adult acute pain.                                                                                  \\
        \cmidrule{2-4}
                                            & max\_initial\_days\_minor        & Integer            & Maximum initial prescription day for minor acute pain.                                                                                  \\
        \cmidrule{2-4}
                                            & mme\_daily\_limit                & Integer            & Daily dose limitation based on morphine milligram equivalence.                                                                          \\
        \midrule
        \multirow{8}{*}{\shortstack{Recovery\\Support}}        
                                            & establish\_program               & Binary             & If a new recovery support program is established.                                                                                       \\
        \cmidrule{2-4}
                                            & expand\_program                  & Binary             & If an existing recovery support program is expanded.                                                                                    \\
        \cmidrule{2-4}
                                            & general\_funding                 & Binary             & If funding is provided to support recovery services.                                                                                    \\
        \cmidrule{2-4}
                                            & dedicated\_funding               & Binary             & If funding is provided specifically for a program.                                                                                      \\
        \cmidrule{2-4}
                                            & certification\_requirement       & Binary             & If certification is required for recovery support organizations.                                                                        \\
        \cmidrule{2-4}
                                            & operating\_standards             & Binary             & If uniform operating standards are set for recovery support.                                                                            \\
        \cmidrule{2-4}
                                            & reporting\_requirement           & Binary             & If data collection and reporting is required.                                                                                           \\
        \cmidrule{2-4}
                                            & target\_population               & Categorical        & If the policy targets specific population, including recovery residents, youth, homeless, incarcerated or detained, and mental illness. \\
        \bottomrule
    \end{tabularx}%
    }
    \label{tab:policycode}%
\end{table*}%

\noindent\textbf{Main Results.}
Table \ref{tab:main_result} summarizes the performance \method in standard forecasting and cross-state generalization. \method consistently achieves the best or sub-best performance across all the evaluation metrics in both settings. Notably, the performance gains are more significant in long-horizon forecasting and cross-state generalization. We attribute such phenomenon to structured policy representation and the explicit modeling of policy-conditioned dynamics. Policy vocabulary can encourage the model to learn intervention-level transition mechanisms rather than state-specific correlations, which is essential for both long-term forecasting and generalization to unseen states.

\vspace{3pt}
\noindent\textbf{Policy Analysis.}
As shown in Table \ref{tab:ablation}, we remove VQ codebook (w/o VQ), relational neighborhood aggregation (w/o KG encoder) and all policy-related modules (w/o Policy) respectively. To further demonstrate the superiority of policy knowledge graph, we replace policy representation with a set of pre-defined policy codes (w/o KG), whose definitions are summarized in Table ~\ref{tab:policycode}. We adopt two categories of policy code: Prescription Control and Recovery Support. Prescription Control policies encompass regulatory measures such as mandatory Prescription Drug Monitoring Program (PDMP) use for prescribers and dispensers, the scope of monitored substances (ranging from Schedules II-V to Drugs of Concern), and quantitative restrictions including maximum initial prescription days for adults and minors, as well as daily dose limits based on morphine milligram equivalence (MME). Recovery Support policies capture programmatic interventions, including the establishment or expansion of recovery programs, funding mechanisms (both general and dedicated), organizational requirements such as certification and operating standards, reporting obligations, and targeted population specifications (e.g., recovery residents, youth, homeless individuals, incarcerated or detained persons, and those with mental illness). The results show that policy knowledge graph and VQ-based policy representation significantly contribute to comprehensive policy modeling. Over-simplifying policies into policy codes may even lead to performance degradation. Pathway-centric intervention encoding can model intervention mechanisms more accurately.

\vspace{3pt}
\noindent\textbf{Feature Analysis.}
As shown in Table \ref{tab:ablation}, we remove economic features (w/o Economic), crime features (w/o Crime) and demographic features (w/o Demographic) respectively to measure the correlation between these features and the opioid epidemic. All the selected features are strongly associated with opioid epidemic. However, removing economic and crime features results in relatively smaller performance drops, suggesting that their contributions are more context-dependent and potentially mediated by policy conditions.

\vspace{3pt}
\noindent\textbf{World Model Analysis.}
As shown in Table \ref{tab:ablation}, we first remove positional encoding from Transformer input (w/o PE), and then replace the GNN encoder with GCN~\cite{kipf2017gcn} and GAT~\cite{velivckovic2018graph} respectively. We also test a two-layer GraphSAGE encoder. The results demonstrate that 1-layer GraphSAGE with positional encoding (default setting) is optimal for our task. Explicit modeling of temporal information is necessary for policy-conditioned dynamic modeling. Effective modeling of opioid dynamics requires a careful balance between spatial diffusion, policy effect and representation stability. GCN has strong neighborhood inductive bias, while attention-based fusion may weaken the policy effects.

\subsection{Opioid Overdose Counterfactual Reasoning}
\label{sec:counterfact}

\begin{figure}[!t]
  \centering
  \subfigure[Policies used in counterfactual reasoning.]{
    \resizebox{0.95\linewidth}{!}{
      \setlength{\tabcolsep}{4pt}
      \begin{tabular}{l p{4.5cm} p{5.5cm}}
        \toprule
        \textbf{Policy ID (Year)} & \textbf{Policy Name}            & \textbf{Topics}                                                     \\
        \midrule
        \rowcolor{gray!5}
        KY H 248 (2023)           & Recovery Housing                & Residential and Hospital-Based Treatment, Recovery Support Services \\
        TN H 215 (2022)           & Drug and Alcohol Rehabilitation & Medication Assisted Treatment, Recovery Support Services            \\
        \bottomrule
      \end{tabular}
    }
    \label{fig:counterfact_policy}
  }

  \subfigure[Opioid overdose death trajectory in Tennessee.]{
    \includegraphics[width=\linewidth]{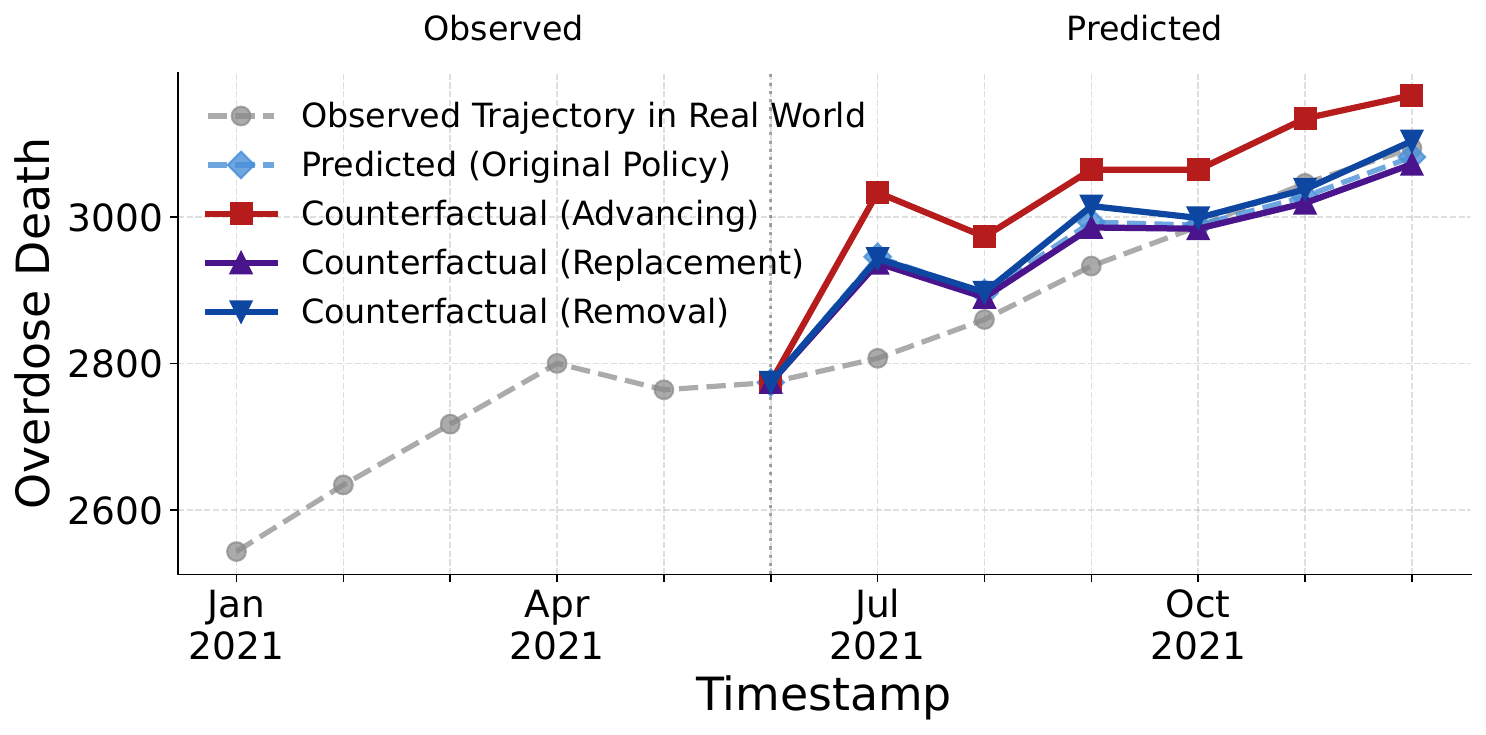}
    \label{fig:counterfact_reasoning}
  }
  \vspace{-10pt}
  \caption{Counterfactual reasoning analysis in Tennessee. (a) Policies used. (b) Predicted overdose death trajectories. }
  \label{fig:counterfactual}
  \vspace{-10pt}
\end{figure}

To demonstrate \method's ability in assessing the effect of alternative policy implementation strategies, we conduct qualitative case study on Tennessee, which has consistently experienced severe opioid overdose burdens~\cite{cdc2023}.
During 2019-2024, Tennessee enacted one policy (TN H 215) about the reformation of Drug Control Policy Office in Apr. 2021.
Detailed introduction of policies used in this study is provided in Table \ref{fig:counterfact_policy}.
In this study, we use the trained \method to predict how opioid epidemic will evolve in Tennessee under following three settings: (1) Advancing: advance the implementation time of the reformation policy by six month; (2) Replacement: replace the reformation policy with recovery housing policy (KY H 248), meanwhile keep the implementation timeline unchanged; (3) Removal: remove policy TN H 215 in Tennessee. The input window is set from Jan. 2021 to Jun. 2021, and we aim to predict the opioid epidemic evolution from Jul. 2021 to Dec. 2021. Figure \ref{fig:counterfactual} summarizes our predicted opioid epidemic trend under ground truth policy trajectory and the three aforementioned settings. The following observations can be obtained: (1) Despite the non-negligible prediction error, our \method correctly simulated the trend of opioid crisis. (2) Advancing the implementation time of TN H 215 can significantly amplify the opioid overdose mortality. TN H 215 aims to strengthen the regulation, licensing, and oversight of substance use disorder treatment providers and recovery residences, whose aggressive implementation could cause systematic substitution. (3) Recovery service expansion is proved to be effective nationwide~\cite{kravitz2020association}. Although the replacement with KY H 248 brings counterfactual effect $\Delta_i(t^*, a')=-45.51$, it is not significantly beneficial. We attribute it to the mismatch between local socio-economic context and detailed policy instruments. (4) Removing TN H 215 results in a slight increase in the predicted opioid overdose death with counterfactual effect $\Delta_i(t^*, a')=61.67$, indicating the effectiveness of recovery service regulation policy.

Despite the promising counterfactual prediction capability of PolicyOOD, the counterfactual effect $\Delta_i(t^*, a')$ reported here can not replace causally identified treatment effects. Policy4OOD does not control for detailed confounding factors or enforce parallel trends assumptions, and the background socioeconomic dynamics remain fixed. Therefore, the analysis should not be interpreted as rigorous causal inference. Besides, counterfactual outcomes are inherently unobservable, meaning the trajectories shown for Advancing, Replacement, and Removal scenarios cannot be directly validated against ground truth. Instead, the model's credibility in qualitative counterfactual reasoning rests on its factual forecasting accuracy.

\begin{figure}[!t]
  \centering
  \subfigure[Candidate policy pool for policy optimization.]{
    \resizebox{0.95\linewidth}{!}{
      \setlength{\tabcolsep}{4pt}
      \begin{tabular}{l p{4.5cm} p{5.5cm}}
        \toprule
        \textbf{Policy ID (Year)} & \textbf{Policy Name}                                   & \textbf{Topics}                                                                                    \\
        \midrule
        \rowcolor{gray!5}
        KY H 248 (2023)           & Recovery Housing                                       & Residential and Hospital-Based Treatment, Recovery Support Services                                \\
        NC S 594 (2022)           & Medicaid Administration                                & Medication Assisted Treatment, Recovery Support Services, Medicaid                                 \\
        \rowcolor{gray!5}
        PA H 1421 (2022)          & Commonwealth Budget                                    & Medication Assisted Treatment, Residential and Hospital-Based Treatment, Recovery Support Services \\
        WV H 3306 (2023)          & Organizational Structure of Drug Control Policy Office & Residential and Hospital-Based Treatment, Recovery Support Services                                \\
        \rowcolor{gray!5}
        RI S 139 (2020)           & Comprehensive Discharge Planning                       & Residential and Hospital-Based Treatment, Recovery Support Services, Other                         \\
        TN H 215 (2022)           & Drug and Alcohol Rehabilitation                        & Medication Assisted Treatment, Recovery Support Services                                           \\
        \rowcolor{gray!5}
        VA S 846 (2023)           & Peer Recovery Specialists                              & Recovery Support Services, Other                                                                   \\
        VA H 277 (2023)           & Certified Recovery Residences                          & Residential and Hospital-Based Treatment, Recovery Support Services                                \\
        \bottomrule
      \end{tabular}
    }
    \label{fig:policy_pool}
  }


  \subfigure[Policy selection on Tennessee.]{
    \centering
    \small
    \scalebox{0.85}{
      \begin{tabular*}{0.5\linewidth}{l l}
        \toprule
        \textbf{Mode}& \textbf{Policy} \\
        \midrule
        Original & TN H 215        \\
        Selected & TN H 215        \\
        \bottomrule
      \end{tabular*}
    }
    \label{fig:policy_select_tn}
  }
  \hspace{5pt}
  \subfigure[Policy selection on Virginia.]{
    \centering
    \small
    \scalebox{0.85}{
      \begin{tabular*}{0.5\linewidth}{l l}
        \toprule
        \textbf{Mode} & \textbf{Policy}      \\
        \midrule
        Original & VA S 846, VA H 277   \\
        Selected & PA H 1421, WV H 3306 \\
        \bottomrule
      \end{tabular*}
    }
    \label{fig:policy_select_va}
  }


  \subfigure[Opioid overdose death trajectory over searched policies on Tennessee.]{
    \includegraphics[width=0.47\linewidth]{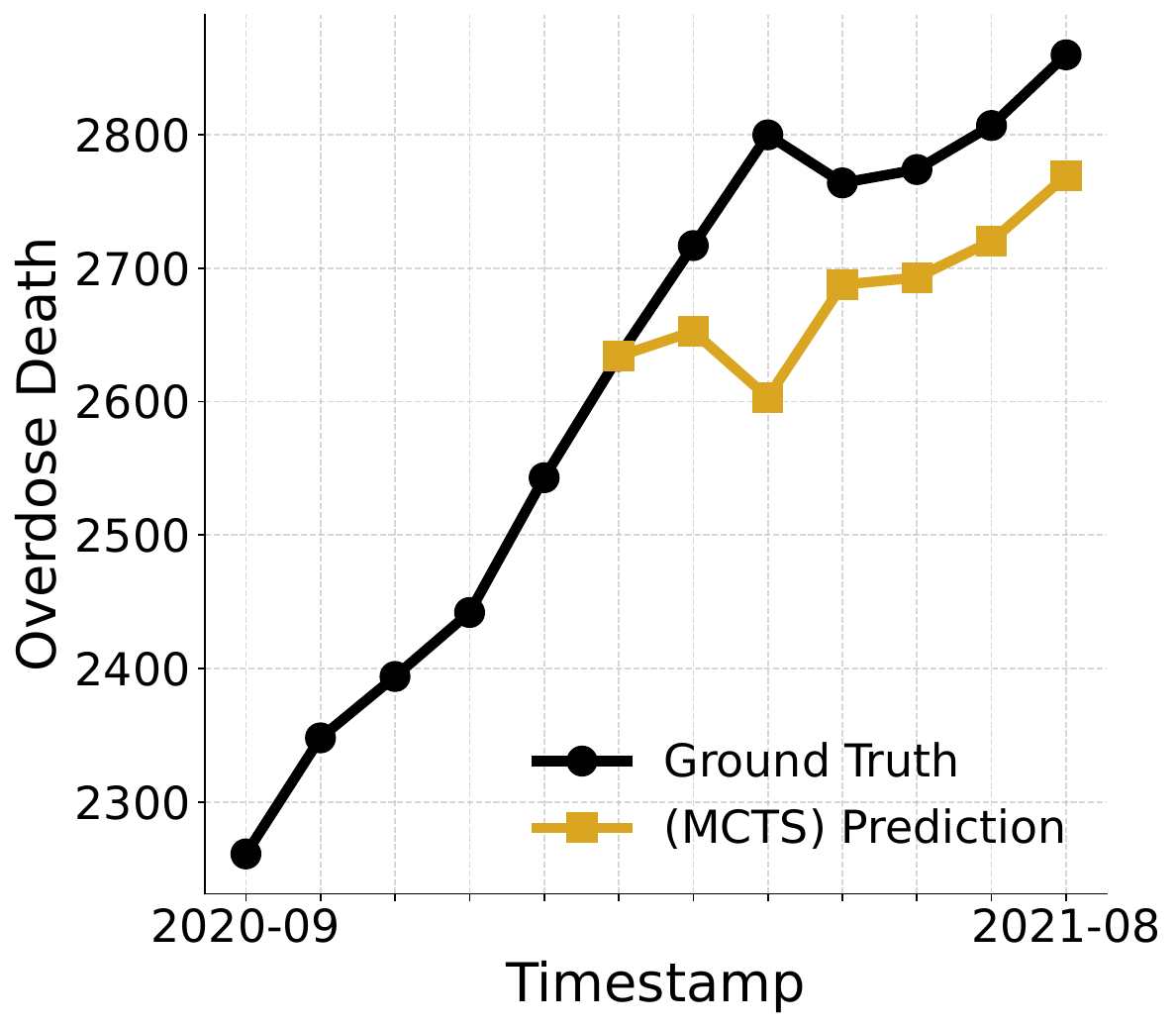}
    \label{fig:policy_traj_tn}
  }
  \hfill
  \subfigure[Opioid overdose death trajectory over searched policies on Virginia.]{
    \includegraphics[width=0.47\linewidth]{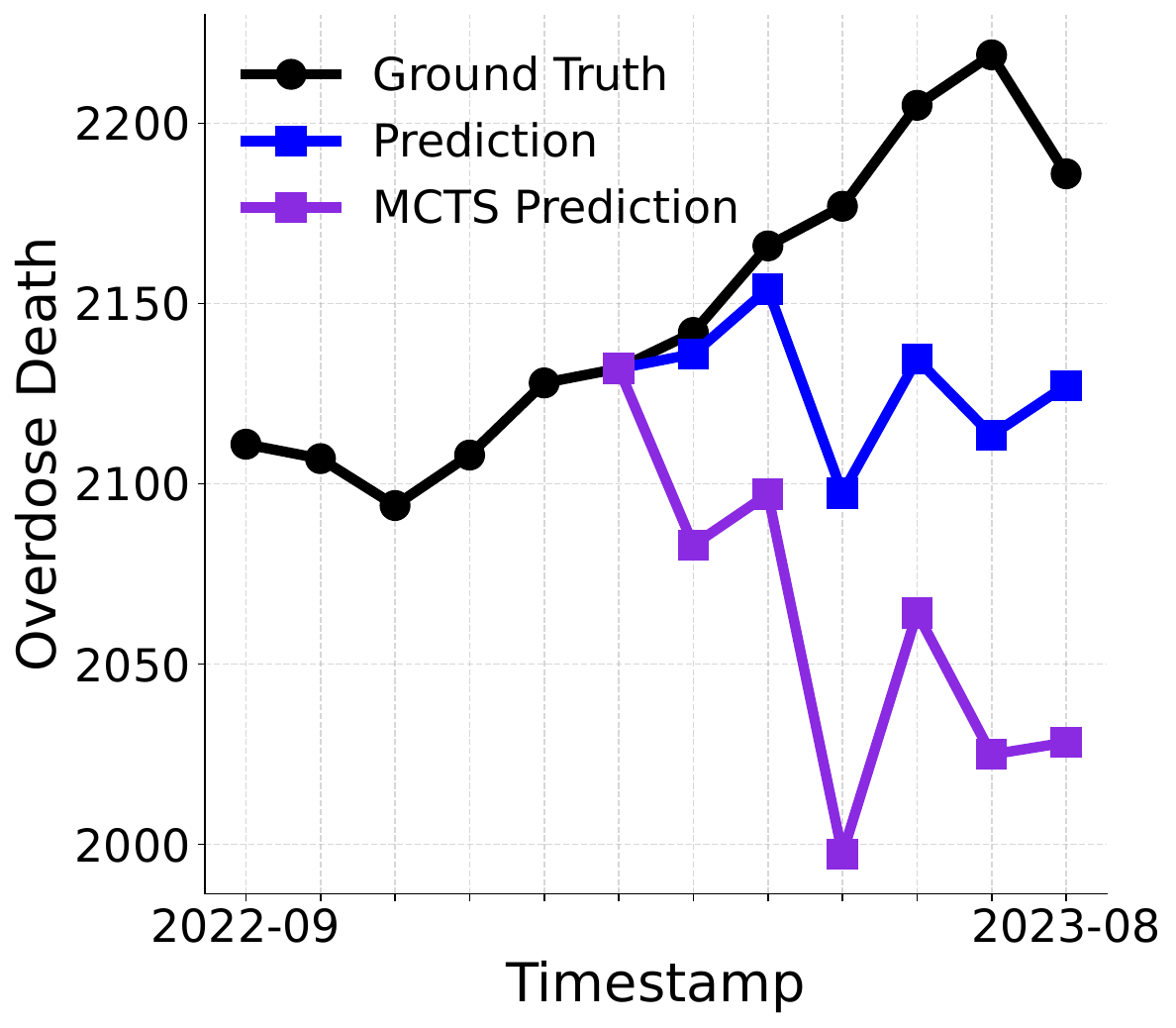}
    \label{fig:policy_traj_va}
  }

  \vspace{-10pt}
  \caption{Case study for policy optimization in Tennessee and Virginia. (a) Candidate policy pool. (b-c) Original and MCTS-selected policies for each state. (d-e) Predicted opioid overdose death trajectories under selected policies.}
  \label{fig:policy_op}
  \vspace{-10pt}
\end{figure}

\subsection{Opioid Overdose Policy Optimization}

Beyond evaluating alternative historical decisions, we further investigate \method's capability in prospective policy optimization. Treating the trained world model as a fast simulator, we employ MCTS to identify candidate policy combinations that minimize predicted cumulative overdose deaths over the forecasting horizon. To construct the candidate policy pool, we categorize available policies into five domains---recovery service regulation, health insurance coverage, funding allocation, governance oversight, and clinical practice---and select a representative policy from each. Detailed policy descriptions are provided in Table~\ref{fig:policy_pool}.
We conduct case studies on Tennessee and Virginia, two states that have also consistently experienced severe opioid overdose burdens. For each state, we augment the base policy pool with its own enacted policy documents to form a state-specific candidate set. Due to differences in the timing of policy enactment, we adopt state-specific forecasting windows: for Tennessee, the input window spans September 2020 to March 2021; for Virginia, it covers September 2022 to March 2023. The embeddings of the selected policies are incorporated into the final token of the Transformer input sequence, conditioning the model on the planned intervention strategy.
As shown in Figure~\ref{fig:policy_op}, \method successfully recovers the policy that was actually enacted in Tennessee during the subsequent months. For Virginia, although the optimized policy set does not coincide with the policies that were actually implemented, the resulting predicted trajectory yields lower cumulative overdose deaths, suggesting that \method can identify potentially more effective policy combinations beyond those historically adopted.

the two case studies together reveal several insights. First, comparing the optimization result in Tennessee with the counterfactual analysis in Section~\ref{sec:counterfact} yields an internally consistent narrative: while MCTS identifies TN H 215 as the near-optimal policy for Tennessee, Section~\ref{sec:counterfact} simultaneously shows that premature implementation of the same policy amplifies overdose mortality. Taken together, these findings suggest that policy effectiveness is jointly determined by \textit{what} is enacted and \textit{when} it is enacted, which is hard to learn for traditional policy evaluation frameworks. Second, for Virginia, the divergence between MCTS-selected policies (PA H 1421 and WV H 3306, both systemic interventions targeting budget allocation and governance infrastructure) and historically enacted policies (VA S 846 and VA H 277, both direct service delivery expansions) suggests that Policy4OOD identifies a latent gap: systemic investment in administrative capacity and sustainable funding may generate more durable reductions in overdose mortality than equivalent investment in direct service provision. This distinction between systematic and service-level interventions aligns with findings in the public health literature~\cite{cerda2025role} and illustrates how the optimization module can surface actionable hypotheses for policy design beyond what historical data directly reveals. Finally, the fact that MCTS recovers a two-policy combination for Virginia rather than a single optimal policy, indicating that the interaction between funding allocation and governance restructuring carries synergistic value that single-policy evaluation methods would miss.
\section{Conclusion}

We present \method, which reframes opioid policy intervention simulation as a world modeling problem. By organizing the design around three questions---\textit{what} policies prescribe, \textit{where} effects manifest, and \textit{when} effects unfold---the framework unifies forecasting, counterfactual reasoning, and policy optimization in a single learned simulator.
Empirically, \method achieves consistently strong forecasting accuracy, with especially pronounced gains in long-horizon prediction and cross-state generalization.
Case studies on qualitative counterfactual reasoning and MCTS-based policy optimization further demonstrate the practical utility of the learned simulator for retrospective policy evaluation and prospective intervention planning.
Future work includes expanding the dataset in temporal coverage and granularity, extending beyond overdose mortality to broader opioid-related outcomes, and validating the counterfactual and policy optimization modules with domain experts.


\bibliographystyle{ACM-Reference-Format}
\bibliography{reference}

@inproceedings{fan2018automatic,
  title     = {Automatic Opioid User Detection from Twitter: Transductive Ensemble Built on Different Meta-graph Based Similarities over Heterogeneous Information Network.},
  author    = {Fan, Yujie and Zhang, Yiming and Ye, Yanfang and Li, Xin},
  booktitle = {IJCAI},
  pages     = {3357--3363},
  year      = {2018}
}

@article{ju2022multi,
  title={Multi-task self-supervised graph neural networks enable stronger task generalization},
  author={Ju, Mingxuan and Zhao, Tong and Wen, Qianlong and Yu, Wenhao and Shah, Neil and Ye, Yanfang and Zhang, Chuxu},
  journal={arXiv preprint arXiv:2210.02016},
  year={2022}
}

@inproceedings{fan2017social,
  title={Social media for opioid addiction epidemiology: Automatic detection of opioid addicts from twitter and case studies},
  author={Fan, Yujie and Zhang, Yiming and Ye, Yanfang and Li, Xin and Zheng, Wanhong},
  booktitle={Proceedings of the 2017 ACM on Conference on Information and Knowledge Management},
  pages={1259--1267},
  year={2017}
}

@article{wang2024gft,
  title={Gft: Graph foundation model with transferable tree vocabulary},
  author={Wang, Zehong and Zhang, Zheyuan and Chawla, Nitesh and Zhang, Chuxu and Ye, Yanfang},
  journal={Advances in Neural Information Processing Systems},
  volume={37},
  pages={107403--107443},
  year={2024}
}

@article{qian2021distilling,
  title={Distilling meta knowledge on heterogeneous graph for illicit drug trafficker detection on social media},
  author={Qian, Yiyue and Zhang, Yiming and Ye, Yanfang and Zhang, Chuxu and others},
  journal={Advances in Neural Information Processing Systems},
  volume={34},
  pages={26911--26923},
  year={2021}
}

@article{ju2022grape,
  title={Grape: Knowledge graph enhanced passage reader for open-domain question answering},
  author={Ju, Mingxuan and Yu, Wenhao and Zhao, Tong and Zhang, Chuxu and Ye, Yanfang},
  journal={arXiv preprint arXiv:2210.02933},
  year={2022}
}

@inproceedings{zhao2023self,
  title={Self-supervised graph structure refinement for graph neural networks},
  author={Zhao, Jianan and Wen, Qianlong and Ju, Mingxuan and Zhang, Chuxu and Ye, Yanfang},
  booktitle={Proceedings of the sixteenth ACM international conference on web search and data mining},
  pages={159--167},
  year={2023}
}

@article{qian2022co,
  title={Co-modality graph contrastive learning for imbalanced node classification},
  author={Qian, Yiyue and Zhang, Chunhui and Zhang, Yiming and Wen, Qianlong and Ye, Yanfang and Zhang, Chuxu},
  journal={Advances in Neural Information Processing Systems},
  volume={35},
  pages={15862--15874},
  year={2022}
}

@inproceedings{oord2017neural,
  title     = {Neural Discrete Representation Learning},
  author    = {van den Oord, Aaron and Vinyals, Oriol and Kavukcuoglu, Koray},
  booktitle = {Advances in Neural Information Processing Systems},
  volume    = {30},
  year      = {2017}
}

@inproceedings{schlichtkrull2018modeling,
  title     = {Modeling Relational Data with Graph Convolutional Networks},
  author    = {Schlichtkrull, Michael and Kipf, Thomas N and Bloem, Peter and van den Berg, Rianne and Titov, Ivan and Welling, Max},
  booktitle = {European Semantic Web Conference},
  pages     = {593--607},
  year      = {2018},
  publisher = {Springer}
}

@article{dellavigna2025policy,
  title     = {Policy diffusion and polarization across US states},
  author    = {DellaVigna, Stefano and Kim, Woojin},
  journal   = {Review of Economic Studies},
  pages     = {rdaf070},
  year      = {2025},
  publisher = {Oxford University Press UK}
}

@inproceedings{stgcn,
  title     = {Spatio-Temporal Graph Convolutional Networks: A Deep Learning Framework for Traffic Forecasting},
  author    = {Bing Yu and Haoteng Yin and Zhanxing Zhu},
  booktitle = {Proceedings of the Twenty-Seventh International Joint Conference on
               Artificial Intelligence, {IJCAI-18}},
  pages     = {3634--3640},
  year      = {2018},
  month     = {7}
}

@article{vaswani2017attention,
  title   = {Attention is all you need},
  author  = {Vaswani, Ashish and Shazeer, Noam and Parmar, Niki and Uszkoreit, Jakob and Jones, Llion and Gomez, Aidan N and Kaiser, {\L}ukasz and Polosukhin, Illia},
  journal = {Advances in neural information processing systems},
  volume  = {30},
  year    = {2017}
}

@article{hamilton2017inductive,
  title   = {Inductive representation learning on large graphs},
  author  = {Hamilton, Will and Ying, Zhitao and Leskovec, Jure},
  journal = {Advances in neural information processing systems},
  volume  = {30},
  year    = {2017}
}

@inproceedings{kipf2017gcn,
  title     = {Semi-Supervised Classification with Graph Convolutional Networks},
  author    = {Thomas N. Kipf and Max Welling},
  booktitle = {International Conference on Learning Representations},
  year      = {2017}
}

@inproceedings{velivckovic2018graph,
  title     = {Graph Attention Networks},
  author    = {Veli{\v{c}}kovi{\'c}, Petar and Cucurull, Guillem and Casanova, Arantxa and Romero, Adriana and Li{\`o}, Pietro and Bengio, Yoshua},
  booktitle = {International Conference on Learning Representations},
  year      = {2018}
}

@article{wen2019prescription,
  title   = {Prescription drug monitoring program mandates: impact on opioid prescribing and related hospital use},
  author  = {Wen, Hefei and Hockenberry, Jason M and Jeng, Philip J and Bao, Yuhua},
  journal = {Health Affairs},
  volume  = {38},
  number  = {9},
  pages   = {1550--1556},
  year    = {2019}
}

@inproceedings{liu2024review,
  title     = {A review of graph neural networks in epidemic modeling},
  author    = {Liu, Zewen and Wan, Guancheng and Prakash, B Aditya and Lau, Max SY and Jin, Wei},
  booktitle = {Proceedings of the 30th ACM SIGKDD conference on knowledge discovery and data mining},
  pages     = {6577--6587},
  year      = {2024}
}

@article{han2025epidemiology,
  title     = {Epidemiology-informed Spatiotemporal Graph Neural Network for heterogeneity-driven interpretable epidemic forecasting},
  author    = {Han, Shuai and Stelz, Lukas and Sokolowski, Thomas R and Zhou, Kai and St{\"o}cker, Horst},
  journal   = {Engineering Applications of Artificial Intelligence},
  volume    = {162},
  pages     = {112764},
  year      = {2025},
  publisher = {Elsevier}
}

@misc{cdc2025,
  title        = {Understanding the Opioid Overdose Epidemic},
  author       = {{Centers for Disease Control and Prevention}},
  year         = {2025},
  note         = {Approximately 105{,}000 drug overdose deaths in 2023 and nearly 80{,}000 involving opioids},
  howpublished = {\url{https://www.cdc.gov/overdose-prevention/about/understanding-the-opioid-overdose-epidemic.html}},
  organization = {CDC}
}

@misc{cdc2023,
  title        = {Health E-Stat 101: Changes in Drug Overdose Mortality and Selected Drug Type by State: United States, 2022 to 2023},
  author       = {{Centers for Disease Control and Prevention}},
  year         = {2025},
  howpublished = {\url{https://www.cdc.gov/nchs/data/hestat/drug-overdose/drug-overdose-2022-2023.htm}},
  organization = {CDC}
}

@misc{cdc2017,
  title        = {Economics of Injury and Violence Prevention},
  author       = {{Centers for Disease Control and Prevention}},
  year         = {2024},
  howpublished = {\url{https://www.cdc.gov/injury-violence-prevention/economics/index.html}},
  organization = {CDC}
}

@article{krueger2017have,
  title   = {Where have all the workers gone? An inquiry into the decline of the US labor force participation rate},
  author  = {Krueger, Alan B},
  journal = {Brookings papers on economic activity},
  volume  = {2017},
  number  = {2},
  pages   = {1},
  year    = {2017}
}

@article{phillips2017pain,
  title     = {Pain management and the opioid epidemic: balancing societal and individual benefits and risks of prescription opioid use},
  author    = {Phillips, Jonathan K and Ford, Morgan A and Bonnie, Richard J},
  year      = {2017},
  publisher = {National Academies Press}
}

@misc{cdc_overdose_stats,
  title        = {FastStats: Drug Overdose Deaths},
  author       = {{Centers for Disease Control and Prevention}},
  year         = {2025},
  note         = {Number of drug overdose deaths and opioid involvement in 2023},
  howpublished = {\url{https://www.cdc.gov/nchs/fastats/drug-overdoses.htm}}
}

@article{salinas2020deepar,
  title     = {DeepAR: Probabilistic forecasting with autoregressive recurrent networks},
  author    = {Salinas, David and Flunkert, Valentin and Gasthaus, Jan and Januschowski, Tim},
  journal   = {International journal of forecasting},
  volume    = {36},
  number    = {3},
  pages     = {1181--1191},
  year      = {2020},
  publisher = {Elsevier}
}

@article{lim2021temporal,
  title     = {Temporal fusion transformers for interpretable multi-horizon time series forecasting},
  author    = {Lim, Bryan and Ar{\i}k, Sercan {\"O} and Loeff, Nicolas and Pfister, Tomas},
  journal   = {International journal of forecasting},
  volume    = {37},
  number    = {4},
  pages     = {1748--1764},
  year      = {2021},
  publisher = {Elsevier}
}

@inproceedings{zhou2021informer,
  title     = {Informer: Beyond efficient transformer for long sequence time-series forecasting},
  author    = {Zhou, Haoyi and Zhang, Shanghang and Peng, Jieqi and Zhang, Shuai and Li, Jianxin and Xiong, Hui and Zhang, Wancai},
  booktitle = {Proceedings of the AAAI conference on artificial intelligence},
  volume    = {35},
  number    = {12},
  pages     = {11106--11115},
  year      = {2021}
}

@article{wu2021autoformer,
  title   = {Autoformer: Decomposition transformers with auto-correlation for long-term series forecasting},
  author  = {Wu, Haixu and Xu, Jiehui and Wang, Jianmin and Long, Mingsheng},
  journal = {Advances in neural information processing systems},
  volume  = {34},
  pages   = {22419--22430},
  year    = {2021}
}

@article{nie2022time,
  title   = {A Time Series is Worth 64Words: Long-term Forecasting with Transformers},
  author  = {Nie, Y},
  journal = {arXiv preprint arXiv:2211.14730},
  year    = {2022}
}

@article{li2017diffusion,
  title   = {Diffusion convolutional recurrent neural network: Data-driven traffic forecasting},
  author  = {Li, Yaguang and Yu, Rose and Shahabi, Cyrus and Liu, Yan},
  journal = {arXiv preprint arXiv:1707.01926},
  year    = {2017}
}

@article{yu2017spatio,
  title   = {Spatio-temporal graph convolutional networks: A deep learning framework for traffic forecasting},
  author  = {Yu, Bing and Yin, Haoteng and Zhu, Zhanxing},
  journal = {arXiv preprint arXiv:1709.04875},
  year    = {2017}
}

@article{wu2019graph,
  title   = {Graph wavenet for deep spatial-temporal graph modeling},
  author  = {Wu, Zonghan and Pan, Shirui and Long, Guodong and Jiang, Jing and Zhang, Chengqi},
  journal = {arXiv preprint arXiv:1906.00121},
  year    = {2019}
}

@article{bai2020adaptive,
  title   = {Adaptive graph convolutional recurrent network for traffic forecasting},
  author  = {Bai, Lei and Yao, Lina and Li, Can and Wang, Xianzhi and Wang, Can},
  journal = {Advances in neural information processing systems},
  volume  = {33},
  pages   = {17804--17815},
  year    = {2020}
}

@article{lee2021systematic,
  title     = {Systematic evaluation of state policy interventions targeting the US opioid epidemic, 2007-2018},
  author    = {Lee, Byungkyu and Zhao, Wanying and Yang, Kai-Cheng and Ahn, Yong-Yeol and Perry, Brea L},
  journal   = {JAMA network open},
  volume    = {4},
  number    = {2},
  pages     = {e2036687--e2036687},
  year      = {2021},
  publisher = {American Medical Association}
}

@article{cerda2021systematic,
  title     = {A systematic review of simulation models to track and address the opioid crisis},
  author    = {Cerd{\'a}, Magdalena and Jalali, Mohammad S and Hamilton, Ava D and DiGennaro, Catherine and Hyder, Ayaz and Santaella-Tenorio, Julian and Kaur, Navdep and Wang, Christina and Keyes, Katherine M},
  journal   = {Epidemiologic reviews},
  volume    = {43},
  number    = {1},
  pages     = {147--165},
  year      = {2021},
  publisher = {Oxford University Press}
}

@article{lim2022modeling,
  title     = {Modeling the evolution of the US opioid crisis for national policy development},
  author    = {Lim, Tse Yang and Stringfellow, Erin J and Stafford, Celia A and DiGennaro, Catherine and Homer, Jack B and Wakeland, Wayne and Eggers, Sara L and Kazemi, Reza and Glos, Lukas and Ewing, Emily G and others},
  journal   = {Proceedings of the National Academy of Sciences},
  volume    = {119},
  number    = {23},
  pages     = {e2115714119},
  year      = {2022},
  publisher = {National Academy of Sciences}
}

@inproceedings{bobashev2018pain,
  title        = {Pain town, an agent-based model of opioid use trajectories in a small community},
  author       = {Bobashev, Georgiy and Goree, Sam and Frank, Jennifer and Zule, William},
  booktitle    = {International Conference on Social Computing, Behavioral-Cultural Modeling and Prediction and Behavior Representation in Modeling and Simulation},
  pages        = {274--285},
  year         = {2018},
  organization = {Springer}
}

@article{mchugh2015prescription,
  title     = {Prescription drug abuse: from epidemiology to public policy},
  author    = {McHugh, R Kathryn and Nielsen, Suzanne and Weiss, Roger D},
  journal   = {Journal of substance abuse treatment},
  volume    = {48},
  number    = {1},
  pages     = {1--7},
  year      = {2015},
  publisher = {Elsevier}
}

@article{clark2014systematic,
  title     = {A systematic review of community opioid overdose prevention and naloxone distribution programs},
  author    = {Clark, Angela K and Wilder, Christine M and Winstanley, Erin L},
  journal   = {Journal of addiction medicine},
  volume    = {8},
  number    = {3},
  pages     = {153--163},
  year      = {2014},
  publisher = {LWW}
}

@article{cerda2025role,
  title   = {The role of prescription opioid and cannabis supply policies on opioid overdose deaths},
  author  = {Cerd{\'a}, Magdalena and Wheeler-Martin, Katherine and Bruzelius, Emilie and Mauro, Christine M and Crystal, Stephen and Davis, Corey S and Adhikari, Samrachana and Santaella-Tenorio, Julian and Keyes, Katherine M and Rudolph, Kara E and others},
  journal = {American journal of epidemiology},
  volume  = {194},
  number  = {3},
  pages   = {791--801},
  year    = {2025}
}

@inproceedings{ertugrul2019castnet,
  title        = {CASTNet: Community-Attentive Spatio-Temporal Networks for Opioid Overdose Forecasting},
  author       = {Ertugrul, Ali Mert and Lin, Yu-Ru and Taskaya-Temizel, Tugba},
  booktitle    = {Proceedings of Joint European Conference on Machine Learning and Knowledge Discovery in Databases (ECML PKDD)},
  pages        = {432--448},
  year         = {2019},
  organization = {Springer}
}

@inproceedings{wang2024timexer,
  title     = {TimeXer: Empowering Transformers for Time Series Forecasting with Exogenous Variables},
  author    = {Wang, Yuxuan and Wu, Haixu and Dong, Jiaxiang and Liu, Yong and Qiu, Yunzhong and Zhang, Haoran and Wang, Jianmin and Long, Mingsheng},
  booktitle = {Advances in Neural Information Processing Systems},
  year      = {2024}
}

@inproceedings{xia2023cast,
  title     = {Deciphering Spatio-Temporal Graph Forecasting: A Causal Lens and Treatment},
  author    = {Xia, Yutong and Liang, Yuxuan and Wen, Haomin and Liu, Xu and Wang, Kun and Zhou, Zhengyang and Zimmermann, Roger},
  booktitle = {Advances in Neural Information Processing Systems},
  year      = {2023}
}

@article{matero2023trop,
  title   = {Opioid death projections with AI-based forecasts using social media language},
  author  = {Matero, Matthew and Giorgi, Salvatore and Curtis, Brenda and Ungar, Lyle H and Schwartz, H Andrew},
  journal = {npj Digital Medicine},
  volume  = {6},
  number  = {1},
  pages   = {35},
  year    = {2023}
}

@article{heuton2024spatiotemporal,
  title     = {Spatiotemporal Forecasting of Opioid-Related Fatal Overdoses: Towards Best Practices for Modeling and Evaluation},
  author    = {Heuton, Kyle and Kapoor, Jyontika and Shrestha, Shikhar and Stopka, Thomas J and Hughes, Michael C},
  journal   = {American Journal of Epidemiology},
  volume    = {194},
  number    = {6},
  pages     = {1776--1782},
  year      = {2024},
  publisher = {Oxford University Press}
}

@article{shojaati2023abm,
  title     = {Opioid-related harms and care impacts of conventional and AI-based prescription management strategies: insights from leveraging agent-based modeling and machine learning},
  author    = {Shojaati, Narjes and Osgood, Nathaniel D},
  journal   = {Frontiers in Digital Health},
  volume    = {5},
  pages     = {1174845},
  year      = {2023},
  publisher = {Frontiers}
}

@article{jin2024survey,
  title     = {A Survey on Graph Neural Networks for Time Series: Forecasting, Classification, Imputation, and Anomaly Detection},
  author    = {Jin, Ming and Koh, Huan Yee and Wen, Qingsong and Zambon, Daniele and Alippi, Cesare and Webb, Geoffrey I and King, Irwin and Pan, Shirui},
  journal   = {IEEE Transactions on Pattern Analysis and Machine Intelligence},
  year      = {2024},
  publisher = {IEEE}
}

@inproceedings{reimers2019sentence,
  title     = {Sentence-{BERT}: Sentence Embeddings using Siamese {BERT}-Networks},
  author    = {Reimers, Nils and Gurevych, Iryna},
  booktitle = {Proceedings of the 2019 Conference on Empirical Methods in Natural Language Processing and the 9th International Joint Conference on Natural Language Processing (EMNLP-IJCNLP)},
  pages     = {3982--3992},
  year      = {2019}
}

@article{browne2012survey,
  title     = {A Survey of Monte Carlo Tree Search Methods},
  author    = {Browne, Cameron B and Powley, Edward and Whitehouse, Daniel and Lucas, Simon M and Cowling, Peter I and Rohlfshagen, Philipp and Tavener, Stephen and Perez, Diego and Samothrakis, Spyridon and Colton, Simon},
  journal   = {IEEE Transactions on Computational Intelligence and AI in Games},
  volume    = {4},
  number    = {1},
  pages     = {1--43},
  year      = {2012},
  publisher = {IEEE}
}

@inproceedings{kocsis2006bandit,
  title     = {Bandit Based Monte-Carlo Planning},
  author    = {Kocsis, Levente and Szepesv{\'a}ri, Csaba},
  booktitle = {European Conference on Machine Learning},
  pages     = {282--293},
  year      = {2006},
  publisher = {Springer}
}

@article{ha2018world,
  title   = {World Models},
  author  = {Ha, David and Schmidhuber, J{\"u}rgen},
  journal = {arXiv preprint arXiv:1803.10122},
  year    = {2018}
}

@inproceedings{hafner2019dream,
  title     = {Dream to Control: Learning Behaviors by Latent Imagination},
  author    = {Hafner, Danijar and Lillicrap, Timothy and Ba, Jimmy and Norouzi, Mohammad},
  booktitle = {International Conference on Learning Representations},
  year      = {2020}
}

@article{hafner2020dreamerv2,
  title   = {Mastering Atari with Discrete World Models},
  author  = {Hafner, Danijar and Lillicrap, Timothy and Norouzi, Mohammad and Ba, Jimmy},
  journal = {arXiv preprint arXiv:2010.02193},
  year    = {2020}
}

@article{hafner2023dreamerv3,
  title   = {Mastering Diverse Domains through World Models},
  author  = {Hafner, Danijar and Pasukonis, Jurgis and Ba, Jimmy and Lillicrap, Timothy},
  journal = {arXiv preprint arXiv:2301.04104},
  year    = {2023}
}

@article{schrittwieser2020mastering,
  title   = {Mastering Atari, Go, Chess and Shogi by Planning with a Learned Model},
  author  = {Schrittwieser, Julian and Antonoglou, Ioannis and Hubert, Thomas and Simonyan, Karen and Sifre, Laurent and Schmitt, Simon and Guez, Arthur and Lockhart, Edward and Hassabis, Demis and Graepel, Thore and others},
  journal = {Nature},
  volume  = {588},
  number  = {7839},
  pages   = {604--609},
  year    = {2020}
}

@article{yan2021videogpt,
  title   = {Videogpt: Video generation using vq-vae and transformers},
  author  = {Yan, Wilson and Zhang, Yunzhi and Abbeel, Pieter and Srinivas, Aravind},
  journal = {arXiv preprint arXiv:2104.10157},
  year    = {2021}
}

@inproceedings{bruce2024genie,
  title     = {Genie: Generative interactive environments},
  author    = {Bruce, Jake and Dennis, Michael D and Edwards, Ashley and Parker-Holder, Jack and Shi, Yuge and Hughes, Edward and Lai, Matthew and Mavalankar, Aditi and Steigerwald, Richie and Apps, Chris and others},
  booktitle = {Forty-first International Conference on Machine Learning},
  year      = {2024}
}

@article{hu2023gaia,
  title   = {Gaia-1: A generative world model for autonomous driving},
  author  = {Hu, Anthony and Russell, Lloyd and Yeo, Hudson and Murez, Zak and Fedoseev, George and Kendall, Alex and Shotton, Jamie and Corrado, Gianluca},
  journal = {arXiv preprint arXiv:2309.17080},
  year    = {2023}
}

@inproceedings{wu2023daydreamer,
  title        = {{DayDreamer}: World Models for Physical Robot Learning},
  author       = {Wu, Philipp and Escontrela, Alejandro and Hafner, Danijar and Abbeel, Pieter and Goldberg, Ken},
  booktitle    = {Conference on Robot Learning},
  pages        = {2226--2240},
  year         = {2023},
  organization = {PMLR}
}

@article{hochreiter1997lstm,
author = {Hochreiter, Sepp and Schmidhuber, J\"{u}rgen},
title = {Long Short-Term Memory},
year = {1997},
issue_date = {November 15, 1997},
volume = {9},
number = {8},
journal = {Neural Comput.},
month = nov,
pages = {1735–1780},
numpages = {46}
}

@article{bai2018empirical,
	author    = {Shaojie Bai and J. Zico Kolter and Vladlen Koltun},
	title     = {An Empirical Evaluation of Generic Convolutional and Recurrent Networks for Sequence Modeling},
	journal   = {arXiv:1803.01271},
	year      = {2018},
}

@inproceedings{wu2020mtgnn,
author = {Wu, Zonghan and Pan, Shirui and Long, Guodong and Jiang, Jing and Chang, Xiaojun and Zhang, Chengqi},
title = {Connecting the Dots: Multivariate Time Series Forecasting with Graph Neural Networks},
year = {2020},
booktitle = {Proceedings of the 26th ACM SIGKDD International Conference on Knowledge Discovery \& Data Mining},
pages = {753–763},
numpages = {11},
series = {KDD '20}
}

@article{kravitz2020association,
  title={Association of Medicaid expansion with opioid overdose mortality in the United States},
  author={Kravitz-Wirtz, Nicole and Davis, Corey S and Ponicki, William R and Rivera-Aguirre, Ariadne and Marshall, Brandon DL and Martins, Silvia S and Cerd{\'a}, Magdalena},
  journal={JAMA network open},
  volume={3},
  number={1},
  pages={e1919066--e1919066},
  year={2020},
  publisher={American Medical Association}
}

@inproceedings{zhang2024dietodin,
author = {Zhang, Zheyuan and Wang, Zehong and Hou, Shifu and Hall, Evan and Bachman, Landon and White, Jasmine and Galassi, Vincent and Chawla, Nitesh V. and Zhang, Chuxu and Ye, Yanfang},
title = {Diet-ODIN: A Novel Framework for Opioid Misuse Detection with Interpretable Dietary Patterns},
year = {2024},
booktitle = {Proceedings of the 30th ACM SIGKDD Conference on Knowledge Discovery and Data Mining},
pages = {6312–6323},
numpages = {12},
series = {KDD '24}
}

@inproceedings{wen2022disentangle,
author = {Wen, Qianlong and Ouyang, Zhongyu and Zhang, Jianfei and Qian, Yiyue and Ye, Yanfang and Zhang, Chuxu},
title = {Disentangled Dynamic Heterogeneous Graph Learning for Opioid Overdose Prediction},
year = {2022},
booktitle = {Proceedings of the 28th ACM SIGKDD Conference on Knowledge Discovery and Data Mining},
pages = {2009–2019},
numpages = {11},
series = {KDD '22}
}

@inproceedings{zhang2025mopihfrs,
author = {Zhang, Zheyuan and Wang, Zehong and Ma, Tianyi and Taneja, Varun Sameer and Nelson, Sofia and Le, Nhi Ha Lan and Murugesan, Keerthiram and Ju, Mingxuan and Chawla, Nitesh V. and Zhang, Chuxu and Ye, Yanfang},
title = {MOPI-HFRS: A Multi-objective Personalized Health-aware Food Recommendation System with LLM-enhanced Interpretation},
year = {2025},
booktitle = {Proceedings of the 31st ACM SIGKDD Conference on Knowledge Discovery and Data Mining V.1},
pages = {2860–2871},
numpages = {12},
series = {KDD '25}
}

@article{li2025interpretable,
  title={Interpretable Graph-Language Modeling for Detecting Youth Illicit Drug Use},
  author={Li, Yiyang and Wang, Zehong and Yuan, Zhengqing and Zhang, Zheyuan and Murugesan, Keerthiram and Zhang, Chuxu and Ye, Yanfang},
  journal={arXiv preprint arXiv:2510.15961},
  year={2025}
}

@inproceedings{wangbeyond,
  title={Beyond Message Passing: Neural Graph Pattern Machine},
  author={Wang, Zehong and Zhang, Zheyuan and Ma, Tianyi and Chawla, Nitesh V and Zhang, Chuxu and Ye, Yanfang},
  booktitle={Forty-second International Conference on Machine Learning},
  year={2025},
}

@inproceedings{wang2025generative,
  title={Generative Graph Pattern Machine},
  author={Wang, Zehong and Zhang, Zheyuan and Ma, Tianyi and Zhang, Chuxu and Ye, Yanfang},
  booktitle={The Thirty-ninth Annual Conference on Neural Information Processing Systems},
  year={2025},
}

@misc{ma2026tgpm,
      title={Temporal Graph Pattern Machine}, 
      author={Yijun Ma and Zehong Wang and Weixiang Sun and Yanfang Ye},
      year={2026},
      eprint={2601.22454},
      archivePrefix={arXiv},
}

@article{gomes2023trends,
  title={Trends in opioid toxicity--related deaths in the US before and after the start of the COVID-19 pandemic, 2011-2021},
  author={Gomes, Tara and Ledlie, Shaleesa and Tadrous, Mina and Mamdani, Muhammad and Paterson, J Michael and Juurlink, David N},
  journal={JAMA Network Open},
  volume={6},
  number={7},
  pages={e2322303},
  year={2023}
}

@inproceedings{wutimesnet,
  title={TimesNet: Temporal 2D-Variation Modeling for General Time Series Analysis},
  author={Wu, Haixu and Hu, Tengge and Liu, Yong and Zhou, Hang and Wang, Jianmin and Long, Mingsheng},
  year={2023},
  booktitle={The Eleventh International Conference on Learning Representations}
}

@inproceedings{liu2023spatio,
  title={Spatio-temporal adaptive embedding makes vanilla transformer sota for traffic forecasting},
  author={Liu, Hangchen and Dong, Zheng and Jiang, Renhe and Deng, Jiewen and Deng, Jinliang and Chen, Quanjun and Song, Xuan},
  booktitle={Proceedings of the 32nd ACM international conference on information and knowledge management},
  pages={4125--4129},
  year={2023}
}


\end{document}